\documentclass{article}

\usepackage{microtype}
\usepackage{graphicx}

\usepackage{booktabs} 

\usepackage{float} 
\usepackage{caption}
\usepackage{subcaption}

\usepackage[dvipsnames]{xcolor}

\usepackage{alphalph}
\usepackage{hyperref}

\usepackage[accepted]{icml2023}

\usepackage{amsmath}
\usepackage{amssymb}
\usepackage{mathtools}
\usepackage{amsthm}

\usepackage[capitalize,noabbrev]{cleveref}

\theoremstyle{plain}
\newtheorem{theorem}{Theorem}[section]

\newtheorem{lemma}[theorem]{Lemma}
\newtheorem{corollary}[theorem]{Corollary}
\theoremstyle{definition}

\theoremstyle{remark}

\usepackage[textsize=tiny]{todonotes}

\icmltitlerunning{On Many-Actions Policy Gradient}

\begin{document}

\twocolumn[
\icmltitle{On Many-Actions Policy Gradient}



\icmlsetsymbol{equal}{*}

\begin{icmlauthorlist}
\icmlauthor{Michal Nauman}{yyy,sch}
\icmlauthor{Marek Cygan}{yyy,comp}
\end{icmlauthorlist}

\icmlaffiliation{yyy}{Informatics Institute, University of Warsaw}
\icmlaffiliation{comp}{Nomagic}
\icmlaffiliation{sch}{Ideas National Centre for Research and Development}

\icmlcorrespondingauthor{Michal Nauman}{nauman.mic@gmail.com}

\icmlkeywords{Machine Learning, ICML}

\vskip 0.3in
]



\printAffiliationsAndNotice{} 

\begin{abstract}
We study the variance of stochastic policy gradients (SPGs) with many action samples per state. We derive a many-actions optimality condition, which determines when many-actions SPG yields lower variance as compared to a single-action agent with proportionally extended trajectory. We propose Model-Based Many-Actions (MBMA), an approach leveraging dynamics models for many-actions sampling in the context of SPG. MBMA addresses issues associated with existing implementations of many-actions SPG and yields lower bias and comparable variance to SPG estimated from states in model-simulated rollouts. We find that MBMA bias and variance structure matches that predicted by theory. As a result, MBMA achieves improved sample efficiency and higher returns on a range of continuous action environments as compared to model-free, many-actions, and model-based on-policy SPG baselines.
\end{abstract}

\section{Introduction}

Stochastic policy gradient (SPG) is a method of optimizing stochastic policy through gradient ascent in the context of reinforcement learning (RL) \citep{williams1992simple, sutton1999policy, peters2006policy}. When paired with powerful function approximators, SPG-based algorithms have proven to be one of the most effective methods for achieving optimal performance in Markov Decision Processes (MDPs) with unknown transition dynamics \citep{schulman2017proximal}. Unfortunately, the exact calculation of the gradient is unfeasible and thus the objective has to be estimated \citep{sutton1999policy}. Resulting variance is known to impact learning speed, as well as performance of the trained agent \citep{konda1999actor, tucker2018mirage}.

On-policy sample efficiency (ie. the number of environment interactions needed to achieve a certain performance level) is particularly affected by variance, as the gradient must be evaluated over long sequences in order to produce a sufficient quality of the SPG estimate \citep{mnih2016asynchronous}. As such, a variety of methods for SPG variance reduction have been proposed. The most widely used is baseline variance reduction, which has been shown to improve algorithms stability and became indispensable to contemporary SPG implementations \citep{peters2006policy, schulman2015high}. Alternative approaches include Q-value bootstrapping \cite{gu2017q}, reducing the effect of long-horizon stochasticity via small discount \cite{baxter2001infinite}, increasing number of samples via parallel agents \cite{mnih2016asynchronous} or using many-actions estimator \cite{asadi2017mean, kool2019estimating, petit2019all, ciosek2020expected}.

In many-actions SPG (MA), the gradient is calculated using more than one action sample per state, without including the follow-up states of additional actions. The method builds upon conditional Monte-Carlo and yields variance that is smaller or equal to that of single-action SPG given fixed trajectory length \cite{bratley2011guide}. These additional action samples can be drawn with \cite{ciosek2020expected} or without replacement \cite{kool2019estimating} and can be generated through rewinding the environment \cite{schulman2015trust} or using a parametrized Q-value approximator \cite{asadi2017mean}. However, drawing additional action samples from the environment is unacceptable in certain settings, while using a Q-network may introduce bias to the gradient estimate. Furthermore, a diminishing variance reduction effect can be achieved by extending the trajectory. This leads to the following questions: 

\begin{enumerate}
    \item Given fixed trajectory length and cost of sampling actions, is SPG variance more favorable when sampling additional actions or extending the trajectory?
    \item Given that more samples translate to smaller variance, what is the bias associated with simulating such additional samples via neural networks?
\end{enumerate}

\begin{figure*}[ht!]
\begin{center}
    \begin{subfigure}{0.37\linewidth}
    \caption{Expected steps to solve (thousands)}
    \includegraphics[width=\textwidth]{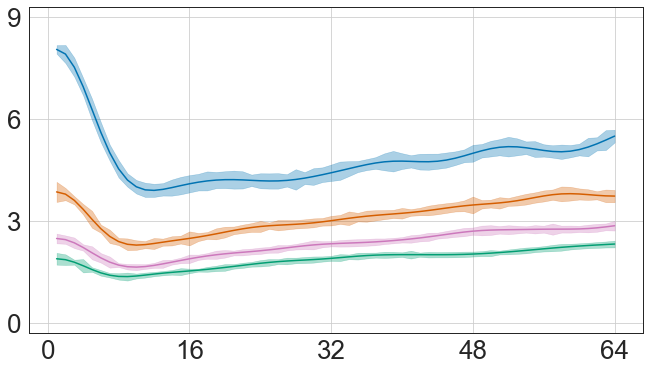}
    \label{fig:cart1}
\end{subfigure}
\hspace{38pt}
\begin{subfigure}{0.37\linewidth}
    \caption{Expected reward gain after update}
    \includegraphics[width=\textwidth]{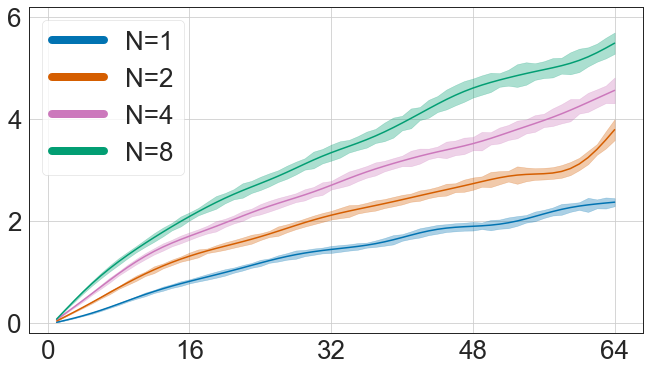}
    \label{fig:cart2}
\end{subfigure}
\vspace{-0.2in}
\caption{Variance reduction leads to better sample efficiency. We train a CartPole Actor-Critic agent with different batch sizes and many action samples per state (denoted as $N$). In Figures \ref{fig:cart1} and \ref{fig:cart2} X-axis denotes batch size (ie. trajectory length) and Y-axis denotes thousands of steps and average performance gain resulting from a single policy update. Increasing batch size leads to better gradient quality at the cost of fewer updates during training. Sampling more actions yields better gradient quality with fewer environment steps.}
\label{fig1}
\end{center}
\vspace{-0.2in}
\end{figure*}

The contributions of this paper are twofold. Firstly, we analyze SPG variance theoretically. We quantify the variance reduction stemming from sampling multiple actions per state as compared to extending the trajectory of a single-action agent. We calculate conditions under which adopting MA estimation leads to greater variance reduction than extending trajectory length. We show that the conditions are often met in RL, but are impossible for contextual bandits. Secondly, we propose an implementation of MA, which we refer to as the Model-Based Many-Actions module (MBMA). The module leverages a learned dynamics model to sample state-action gradients and can be used in conjunction with any on-policy SPG algorithm. MBMA yields a favorable bias/variance structure as compared to learning from states simulated in the dynamics model rollout \citep{janner2019trust, kaiser2019model, hafner2019dream} in the context of on-policy SPG. We validate our approach and show empirically that using MBMA alongside PPO \citep{schulman2017proximal} yields better sample efficiency and higher reward sums on a variety of continuous action environments as compared to many-actions, model-based and model-free PPO baselines. 

\section{Background}

A Markov Decision Process (MDP) \cite{puterman2014markov} is a tuple $(S, A, R, p, \gamma)$, where $S$ is a countable set of states, $A$ is a countable set of actions, $R(s, a)$ is the state-action reward, $p(s'|s, a)$ is a transition kernel (with the initial state distribution denoted as $p_{0}$) and $\gamma \in (0,1]$ is a discount factor. A policy $\pi(a|s)$ is a state-conditioned action distribution. Given a policy $\pi$, MDP becomes a Markov reward process with a transition kernel $p^{\pi}(s' | s) = \int_{a} \pi (a|s) ~ p(s'|s, a) ~ d a$, which we refer to as the underlying Markov chain. The underlying Markov chain is assumed to have finite variance, a unique stationary distribution denoted as $p^{\pi}_{0}$ \cite{ross1996stochastic, konda1999actor}, $t$-step stationary transition kernel $p^{\pi}_{t}$ and a unique discounted stationary distribution denoted as $p^{\pi}_{*}$. Interactions with the MDP according to some policy $\pi$ are called trajectories and are denoted as $\tau^{\pi}_{T} (s_t) = \left((s_t, a_t, r_t), ..., (s_{t+T}, a_{t+T}, r_{t+T})\right)$, where $a_t \sim \pi(a_t|s_t)$, $r_t \sim R(s, a)$ and $s_{t+1} \sim p(s_{t+1}|s_{t}, a_{t})$. The value function $V^{\pi} (s) = \mathrm{E}_{\tau^{\pi}_{\infty} (s)} [\sum_{t=0}^{\infty} \gamma^t R(s_t, a_t)]$ and Q-value function $Q^{\pi} (s, a) = \mathrm{E}_{\tau^{\pi}_{\infty} (s|a)} [\sum_{t=0}^{\infty} \gamma^t R(s_t, a_t)] = R(s, a) ~ + ~ \gamma \mathrm{E}_{s' \sim p(s'| s, a)} [V^{\pi} (s')] $ sample $a_t$ according to some fixed policy $\pi$. State-action advantage is defined as $A^{\pi} (s, a) = Q^{\pi} (s, a) - V^{\pi} (s)$. An optimal policy is a policy that maximizes discounted total return $J = \int_{s_0} V^{\pi} (s_0) ~ds_0$. 

\subsection{On-policy SPG}

Given a policy parametrized by $\theta$, the values of $\theta$ can be updated via SPG $\theta \leftarrow \theta + \nabla_{\theta} J$. Since we are interested only in gradient wrt. $\theta$, we drop it from the gradient notation in further uses. The SPG is given by \cite{sutton2018reinforcement}:

\begin{equation}
\label{eq:1}
\begin{split}
    \nabla J & = \underset{s \sim p^{\pi}_{*}}{\mathbb{E}} ~ \underset{a \sim \pi}{\mathbb{E}} ~ Q^{\pi} (s, a) ~ \nabla \log \pi (a | s)
\end{split}
\end{equation}

As such, SPG is proportional to a double expectation of $Q^{\pi} (s, a)  \nabla_{\theta} \log \pi (a | s)$, with the outer expectation taken wrt. the discounted stationary distribution $p^{\pi}_{*}$ and the inner expectation taken wrt. policy $\pi$. The gradient can be estimated via a trajectory sampled according to the policy \cite{nota2020policy, wu2022understanding}. We denote $\nabla \hat{J}$ as the estimator, $\nabla 
J(s_t, a_t) = Q^{\pi} (s_t, a_t) \nabla \log \pi (a_t | s_t)$ with $s_t, a_t \sim p_{t}^{\pi}, \pi$. Then, SPG can be calculated:

\begin{equation}
\label{eq:2}
\begin{split}
    \nabla \hat{J} = \frac{1}{T} \sum_{t=0}^{T-1} ~ \gamma^{t} ~ \nabla J(s_t, a_t)
\end{split}
\end{equation}

In the setup above, the outer expectation of Equation \ref{eq:1} is estimated via Monte-Carlo \citep{metropolis1949monte} with $T$ state samples drawn from the non-discounted stationary distribution $p^{\pi}_{0}$; and the inner expectation is estimated with a single action per state drawn from the policy $\pi(a|s)$. The resulting variance can be reduced to a certain degree by a control variate, with state value being a popular choice for such baseline \citep{schulman2015high}. Then, the Q-value from Equation \ref{eq:1} is replaced by $A^{\pi} (s_t, a_t)$. If the state value is learned by a parametrized approximator, it is referred to as the \textit{critic}. Critic bootstrapping \citep{gu2017q} is defined as $Q^{\pi}(s, a) = R(s, a) + \gamma V^{\pi}(s')$ with $s' \sim p(s'|s, a)$ and can be used to balance the bias-variance tradeoff of Q-value approximations. 

\subsection{On-policy Many-Actions SPG}

Given a control variate, the variance of policy gradient can be further reduced by approximating the inner integral of Equation \ref{eq:2} with a quadrature of $N>1$ action samples. Then, $\nabla \hat{J}$ is equal to: 

\begin{equation}
\label{eq:sum}
    \nabla \hat{J} = \underbrace{\frac{1}{T} \sum_{t=0}^{T-1} ~ \gamma^{t} ~ \underbrace{\frac{1}{N} \sum_{n=0}^{N-1} \nabla J(s_t, a_{t}^{n})}_\text{$N$ actions per state}}_\text{$T$ state samples in a trajectory}
\end{equation}

Where $a_{t}^{n}$ denotes the $n^{th}$ action sampled at state $s_t$. Furthermore, MDP transitions are conditioned only on the first action performed (ie. $p^{\pi}(s_{t+1}|s_t, a_{t}^{n}) = p^{\pi}(s_{t+1}|s_t) \iff n\neq0$). Implementations of such an approach were called \textit{all-action policy gradient} or \textit{expected policy gradient} \citep{asadi2017mean, petit2019all, ciosek2020expected}. As follows from the law of iterated expectations, the many-actions (MA) estimator is unbiased and yields lower or equal variance as compared to single-action SPG with equal trajectory length \citep{petit2019all}. Since the policy log probabilities are known, using MA requires approximating the Q-values of additional action samples. As such, MA is often implemented by performing rollouts in a rewinded environment \citep{schulman2015trust, kool2019buy, kool2019estimating} or by leveraging a Q-network at the cost of bias \citep{asadi2017mean, petit2019all, ciosek2020expected}. The variance reduction stemming from using MA has been shown to increase both performance and sample efficiency of SPG algorithms \citep{schulman2015trust, kool2019estimating}.

\section{Variance of Stochastic Policy Gradients}

Throughout the section, we assume no stochasticity induced by learning Q-values and we treat Q-values as known. Furthermore, when referring to SPG variance, we refer to the diagonal of the policy parameter variance-covariance matrix. Finally, to unburden the notation, we define $\Upsilon^{t} = \gamma^{t} ~ \nabla J(s_t, a_t)$ and $\bar{\Upsilon}^{t} = \gamma^{t} ~ \mathbb{E}_{a \sim \pi} \nabla J(s_t,a_t)$, where we skip the superscript when $t=0$. Similarly, we use $\mathbb{O}_{a} (\cdot) = \mathbb{O}_{a \sim \pi}(\cdot)$, $\mathbb{O}_{s}(\cdot) = \mathbb{O}_{s \sim p^{\pi}_{0}}(\cdot)$ and $\mathbb{O}_{s, a}(\cdot) = \mathbb{O}_{s_t, a_t \sim p_{t}^{\pi}, \pi}(\cdot)$, where $\mathbb{O}$ denotes expected value, variance and covariance operators. As shown, given fixed trajectory length $T$, MA-SPG variance is smaller or equal to the variance of single-action agent \citet{petit2019all, ciosek2020expected}. However, approximating the inner expectation of SPG always uses resources (ie. compute or environment interactions), which could be used to reduce the SPG variance through other means (eg. extending the trajectory length). To this end, we extend existing results \cite{petit2019all, ciosek2020expected} by comparing the variance reduction stemming from employing MA as opposed to using regular single-action SPG with an extended trajectory length. If the underlying Markov chain is ergodic the variance of SPG, denoted as $\mathbf{V}$, can be calculated via Markov chain Central Limit Theorem \cite{jones2004markov, brooks2011handbook}:

\begin{equation}
\label{eq:3}
\begin{split}
    & \mathbf{V} = \frac{1}{T} ~ \underset{s, a}{\mathrm{Var}}  ~\bigl[ \Upsilon \bigr] + 2 \sum_{t=1}^{T-1} \frac{T-t}{T^2} ~ \underset{s, a}{\mathrm{Cov}} ~ \bigl[ \Upsilon,  \Upsilon^{t} \bigr]
\end{split}
\end{equation}

The states underlying $\Upsilon$ and $\Upsilon^{t}$ are sampled from the undiscounted stationary distribution $p_{0}^{\pi}$ and the t-step stationary transition kernel $p_{t}^{\pi}$ respectively. As follows from the ergodic theorem \cite{norris1998markov}, conditional probability of visiting state $s_t$ given starting in state $s_0$ with action $a_{0}^{0}$ approaches the undiscounted stationary distribution $p^{\pi}_{0}$ exponentially fast as $t$ grows $\lim_{t\to\infty} p(s_t | s_0, a_{0}^{1})= p_{0}^{\pi}(s_t)$. Therefore, $\mathrm{Cov}_{t} \geq \mathrm{Cov}_{t+1}$, as well as  $\lim_{t\to\infty} \mathrm{Cov}_t = 0$. Equation \ref{eq:3} shows the well-known result that increasing the trajectory length $T$ decreases $\mathbf{V}$. This result contextualizes the success of parallel SPG \cite{mnih2016asynchronous}. Unfortunately, the form above assumes single action per state.

\subsection{Variance Decomposition}

To quantify variance reduction stemming from many action samples, we decompose $\mathbf{V}$ into sub-components. We include derivations in Appendix \ref{appendix11}.

\begin{lemma}
\label{lem:lemma1}
Given ergodic MDP, SPG with $N$ action samples per state and $T$ states, $\mathbf{V}$ can be decomposed into:

\begin{equation}
\label{eq:table1}
\begin{split}
    & \underset{s, a}{\mathrm{Var}}  ~\bigl[ \Upsilon \bigr] = \underset{s}{\mathrm{Var}} ~ \bigl[\bar{\Upsilon}\bigr] + \frac{1}{N} ~ \underset{s}{\mathbb{E}} ~ \underset{a}{\mathrm{Var}} ~ \bigl[  \Upsilon \bigr] \\
    & \underset{s, a}{\mathrm{Cov}} ~ \bigl[\Upsilon, \Upsilon^t \bigr] = \underset{s, a}{\mathrm{Cov}} \bigl[ \hat{\Upsilon}, \hat{\Upsilon}^{t}\bigr] + \frac{1}{N} ~ \underset{s}{\mathbb{E}} ~ \underset{s, a}{\mathrm{Cov}} \bigl[  \Upsilon,  \Upsilon^{t} \bigr]
\end{split}
\end{equation}
\end{lemma}

\begin{table*}[t]
\centering
\caption{Decomposed trace of variance-covariance matrix divided by the number of parameters. The components were estimated by marginalizing Q-values, with Equation \ref{eq:sum} and Lemma \ref{lem:lemma1} using $125 000$ non-baselined interactions. The last two columns record the best performance during $500k$ environment steps (average performance shown in the brackets). The performance of SPG variants was measured during $500k$ training steps with additional action samples drawn from the environment. With most variance depending on the policy, MA often yields better performance than single-action agents with extended trajectories. We detail the setting in Appendix \ref{appendix3}.}
\vskip 0.15in
 {\renewcommand{\arraystretch}{1.1}%
\begin{tabular}{||l||c|c|c|c||}
\hline
    & \multicolumn{2}{|c|}{\textsc{Variance component}} & \multicolumn{2}{|c|}{\textsc{Performance}} \\ \hline\hline
    \textsc{Task} & \textsc{Marginalized policy} & \textsc{Policy-dependent} & $(T, \, N) = (1024, \, 2)$ & $(T, \, N) = (2048, \, 1)$  \\
    \hline\hline
    \textsc{ball catch} & $0.026 \, (3\%)$ & $0.819 \, (97\%)$ & $905 \, (708)$ & $920 \, (715)$ \\
    \textsc{cart swingup} & $0.006 \, (1\%)$ &$ 5.736 \, (99\%)$ & $837 \, (670)$ & $801 \, (669)$ \\
    \textsc{cheetah run} & $0.006 \, (1\%)$ & $1.615 \, (99\%)$ & $208 \, (131)$ & $204 \, (126)$ \\
    \textsc{finger spin} & $0.026 \, (18\%)$ & $0.122 \, (82\%)$ & $304 \, (187)$ & $281 \, (179)$ \\
    \textsc{reacher easy} & $2.269 \, (39\%)$ & $3.565 \, (61\%)$ & $428 \, (262)$ & $776 \, (488)$ \\
    \textsc{walker walk} & $0.081 \, (1\%)$ & $11.786 \, (99\%)$ & $509 \, (315)$ & $465 \, (287)$ \\
    \hline
\end{tabular}}
\label{maasdc_grid2}
\vskip -0.1in
\end{table*}

Combining Lemma \ref{lem:lemma1} with Equation \ref{eq:3} yields an expression for decomposed SPG variance, where we group components according to dependence on $N$:

\begin{equation}
\label{eq:varrrr}
\begin{split}
    & T ~ \mathbf{V} = \underbrace{\underset{s}{\mathrm{Var}} ~ \bigl[ \hat{\Upsilon} \bigr] +  2\sum_{t=1}^{T-1} ~ \frac{T-t}{T} ~ \underset{s, a}{\mathrm{Cov}} \bigl[ \hat{\Upsilon}, \hat{\Upsilon}^{t} \bigr]}_\text{Marginalized policy variance}\\
    & + \underbrace{\frac{1}{N} ~ \underset{s}{\mathbb{E}} ~ \Bigl( ~ \underset{a}{\mathrm{Var}} ~ \bigl[  \Upsilon \bigr]
    + 2\sum_{t=1}^{T-1} ~ \frac{T-t}{T} ~ \underset{s, a}{\mathrm{Cov}} \bigl[  \Upsilon,  \Upsilon^{t} \bigr] \Bigr)}_\text{Policy-dependent variance}
\end{split}
\end{equation}

Given $N=1$, the variance simplifies to a single-action case. The statement shows that SPG variance can be decomposed into: marginalized policy variance, which stems from the underlying Markov chain and is decreased only by trajectory length ($T$); and policy-dependent variance, which indeed is reduced by both sampling more actions per state ($N$) and increasing trajectory length ($T$). Table \ref{maasdc_grid2} shows estimated variance components and performance of two SPG estimators ($T=1024; N=2$ and $T=2048; N=1$) for 6 Deepmind Control Suite (DMC) environments. In particular, the table shows that with Q-values marginalized, the policy is responsible for around $90\%$ of SPG variance in tested environments. 

\subsection{Measuring Variance Reduction}

We proceed with the analytical analysis of the variance reduction stemming from increasing $N$ and $T$.

\begin{lemma}
\label{lemma2}
Given ergodic MDP, SPG with $N$ action samples per state and $T$ states, variance reduction stemming from increasing $N$ by $1$ (denoted as $\Delta_N$) and variance reduction stemming from increasing the trajectory length to $T + \delta T$ with $\delta \in (0,\infty)$ (denoted as $\Delta_T$) are equal to:

\begin{equation}
\begin{split}
    & \frac{\Delta_N}{\alpha_N } = \underset{s}{\mathbb{E}} ~ \Bigl(\underset{a}{\mathrm{Var}} ~ \bigl[ \Upsilon \bigr] +2 \sum_{t=1}^{T-1} \frac{T-t}{T} ~ \underset{s, a}{\mathrm{Cov}} \bigl[ \Upsilon, \Upsilon^t \bigr] \Bigr) \\
    & \frac{\Delta_T}{\alpha_T } = \underset{s, a}{\mathrm{Var}} ~ \bigl[\Upsilon\bigr] + 2\sum_{t=1}^{T-1} \Bigl( \frac{T-t}{T} - \frac{t}{T+\delta T} \Bigr) \underset{s, a}{\mathrm{Cov}} \bigl[ \Upsilon, \Upsilon^t \bigr] \\
    & \alpha_N = \frac{-1}{T(N^2 + N)} \quad \textrm{and} \quad \alpha_T = \frac{-\delta}{T + \delta T}
\end{split}
\end{equation}

\end{lemma}

Derivation of Lemma \ref{lemma2} is detailed in Appendix \ref{appendix12}. Lemma \ref{lemma2} shows the diminishing variance reduction stemming from increasing $N$ by $1$ or $T$ by $\delta T$. Incorporating $\delta$ captures the notion of relative costs of increasing $N$ and $T$. If $\delta = 1$, then the cost of increasing $N$ by $1$ (sampling one more action per state in trajectory) is equal to doubling the trajectory length. Now, it follows that many-actions yield better variance reduction than increasing trajectory length only if $\Delta_N \leq \Delta_T$ for given values of $N$, $T$, and $\delta$.

\begin{theorem}
\label{theorem1}
Given ergodic MDP, SPG with $N$ action samples per $T$ states, variance reduction stemming from increasing $N$ by $1$ is bigger than variance reduction stemming from increasing $T$ by $\delta T$ for $\delta = 1$ and $N = 1$ when:

\begin{equation}
\label{theorem1_eqn}
\begin{split}
    & \sum_{t=1}^{T-1} \frac{t}{T} \underset{s, a}{\mathrm{Cov}} \bigl[ \Upsilon, \Upsilon^t \bigr] \geq \underset{s}{\mathrm{Var}} \bigl[\hat{\Upsilon}\bigr] + 2\sum_{t=1}^{T-1} \frac{T - t}{T} \underset{s,a}{\mathrm{Cov}} \bigl[\hat{\Upsilon}, \hat{\Upsilon}^{t} \bigr]
\end{split}
\end{equation}
\end{theorem}

For derivation with $N \geq 1$ and $\delta \in (0,\infty)$ see Equation \ref{eq:theoremsupp} in Appendix \ref{appendix13}. The theorem represents a condition under which optimal to switch from regular SPG (MA-SPG with $N=1$) to MA-SPG with $N=2$. Surprisingly, the optimality condition for $\delta = 1$ and $N=1$ is dependent solely on the covariance structure of the data. As follows from Theorem \ref{theorem1}, MA is optimal when the weighted sum of MDP covariances exceeds the variance of the Markov Chain underlying the MDP. As follows, MA is most effective in problems where action-dependent covariance constitutes a sizeable portion of the total SPG variance (ie. problems where future outcomes largely depend on actions taken in the past and consequently, $\nabla_{\theta}J(s_{t+k}, a_{t+k})$ largely depends upon $a_t$).

\begin{corollary}
\label{cor111}
Given ergodic MDP, SPG with $N$ action samples per state and $T$ states, the SPG variance reduction from increasing $\Delta N = 1$ is bigger than SPG variance reduction from $\Delta T = \delta T$ when:

\begin{equation}
\label{cor1}
    \frac{ \underset{s}{\mathrm{Var}} ~ \bigl[\hat{\Upsilon}\bigr]}{\underset{s}{\mathbb{E}} ~ \underset{a}{\mathrm{Var}} ~ \bigl[\Upsilon\bigr]} \leq
    \frac{1 - \delta N}{\delta (N^2 + N)} 
\end{equation}
\end{corollary}

The corollary above is a specific case of Theorem \ref{theorem1}. By assuming a contextual bandit problem, the covariances are equal to zero and the optimality condition is vastly simplified. As follows from the definition of variance, the LHS of Equation \ref{cor1} is greater or equal to $0$. However, the RHS becomes negative when $\delta N > 1$. Since $N \geq 1$, it follows that MA is never optimal for bandits if $\delta \geq 1$ (ie. the cost of acquiring an additional action sample is equal to or greater than the cost of acquiring an additional state sample). Whereas the efficiency of MA for contextual bandits is restricted, Theorem \ref{theorem1} shows that MA can be a preferable strategy for gradient estimation in MDPs. We leave researching the optimality condition for setting with sampled Q-values or deterministic policy gradients for future work. 

\section{Model-Based Many-Actions SPG}

Given a fixed amount of interactions with the environment, our theoretical analysis is related to two notions in on-policy SPG algorithms: achieving better quality gradients through MA via Q-network (QMA) \cite{asadi2017mean, petit2019all, ciosek2020expected}; and achieving better quality gradients through simulating additional transitions via dynamics model in model-based SPG (MB-SPG) \cite{janner2019trust}. Building on theoretical insights, we propose Model-Based Many-Actions (MBMA), an approach that bridges the two themes described above. MBMA leverages a learned dynamics model in the context of MA-SPG. As such, MBMA allows for MA estimation by calculating Q-values of additional action samples by simulating a critic-bootstrapped trajectory within a dynamics model, consisting of transition and reward networks \cite{ha2018recurrent, hafner2019dream, kaiser2019model, gelada2019deepmdp, schrittwieser2020mastering} which we explain in Appendix \ref{newappendix}. MBMA can be used in conjunction with any on-policy SPG algorithm. 

\subsection{MBMA and MA-SPG}

In contrast to existing implementations of MA-SPG, MBMA does not require Q-network for MA estimation. Using a Q-network to approximate additional action samples yields bias. Whereas the bias can theoretically be reduced to zero, the conditions required for such bias annihilation are unrealistic \cite{petit2019all}. Q-network learns a non-stationary target \cite{van2016deep} that is dependent on the current policy. Furthermore, generating informative samples for multiple actions is challenging given single-action supervision. This results in unstable training when Q-network is used to bootstrap the policy gradient \cite{mnih2015human, van2016deep, gu2017q, haarnoja2018soft}. The advantage of MBMA when compared to QMA is that both reward and transition networks learn stationary targets throughout training, thus offering better convergence properties and lower bias. Such bias reduction comes at the cost of additional computation. Whereas QMA approximates Q-values within a single forward calculation, MBMA sequentially unrolls the dynamics model for a fixed amount of steps.

\subsection{MBMA and MB-SPG}

From the perspective of model-based on-policy SPG, MBMA builds upon on-policy Model-Based Policy Optimization (MPBO) \cite{janner2019trust} but introduces the distinction between two roles for simulated transitions: whereas MBPO calculates gradient at simulated states, we propose to use information from the dynamics model by backpropagating from real states with simulated actions (i.e. simulating Q-values of those actions). As such, we define MBMA as an idea that we do not calculate gradients at simulated states, but instead use the dynamics model to refine the SPG estimator through MA variance reduction. Not calculating gradients at simulated states greatly affects the resulting SPG bias. When backpropagating SPG through simulated states, SPG is biased by two approximates: the Q-value of simulated action; and log-probability calculated at the output of the transition network. The accumulated error of state prediction anchors the gradient on log probabilities which should be associated with different states. MBPO tries to reduce the detrimental effect of compounded dynamics bias by simulating short-horizon trajectories starting from real states. In contrast to that, by calculating gradients at real states, MBMA biases the SPG only through its Q-value approximates, allowing it to omit the effects of biased log probabilities. Such perspective is supported by Lipschitz continuity analysis of approximate MDP models \cite{asadi2018lipschitz, gelada2019deepmdp}. We investigate bias stemming from strategies employed by QMA, MBMA, and MBPO in the table below. In light of the above arguments and our theoretical analysis, we hypothesize that using the dynamics model for MA estimation might yield a more favorable bias-variance tradeoff as compared to using the dynamics model to sample additional states given a fixed simulation budget.

\begin{table}[h]
\caption{SPG per-parameter bias associated with action (MA) and state (MS) sample simulation. $Q$ and $\hat{Q}$ denote the true Q-value and approximate Q-value of a given state-action pair respectively; $s^{*}$ denotes the output of the transition model; and $\mathcal{K}$ denotes the Lipschitz norm of $f_{s} = \nabla \log \pi (a | s)$. For MS the bias is an upper bound. We include extended calculations in Appendix \ref{appendix14}.}
\label{sample-table2}
\begin{center}
{\renewcommand{\arraystretch}{1.2}%
\begin{tabular}{||l||c||}
\hline
 & $\nabla J(s, a) -\nabla \hat{J}(s, a)$ \\
\hline\hline
\textsc{MA} $=$ & $f_{s} (Q - \hat{Q}) $ \\
\textsc{MS} $\leq$ & $f_{s} (Q - \hat{Q}) + \sqrt{(\mathcal{K}(s - \hat{s}))^2 + f_{s}^2 (Q^2 - Q)}$ \\
\hline
\end{tabular}}
\end{center}
\vskip -0.1in
\end{table}

\section{Experiments} 

\subsection{Experimental Setting}

We investigate the effect of bias-variance on the performance of on-policy SPG agents. We compare 4 algorithms implemented with a PPO policy: standard PPO; QMA; MBPO and MBMA. To isolate the effect of bias-variance on agents performance, we implement identical agents that differ only on two dimensions: \textit{which} samples are simulated (ie. no simulation (PPO), state sample simulation (MBPO), action sample simulation (QMA and MBMA)); and \textit{how} samples are simulated (ie. Q-network (QMA) as opposed to dynamics model (MBPO and MBMA)) \textit{ceteris paribus}. We deliberately use the simulated samples only in SPG estimation. As such, the differences in performance stem solely from the bias-variance of specific SPG estimators and the resulting gradient quality. Such an experimental setup reflects the two questions posed in the Introduction:

\begin{enumerate}
    \item By comparing MBPO-PPO and MBMA-PPO we compare variance reduction of many-actions (MBMA) as opposed to extending the trajectory length (MBPO) in the MB-SPG context and validate our theoretical contribution
    \item By comparing QMA-PPO and MBMA-PPO we observe the bias accumulation resulting from simulating action with Q-network (QMA) as opposed to dynamics models (MBMA)
    \item By comparing the bias-free high-variance method (PPO) to biased low-variance methods (QMA, MBPO, and MBMA) we investigate how various levels of bias-variance translate to on-policy SPG performance
\end{enumerate}

Note, that we consider on-policy SPG setting. As such, we pair MBPO with an on-policy PPO agent, as opposed to an off-policy SAC agent considered in the original implementation. Algorithm \ref{alg:algo} shows the implementation of MBMA and MBPO used in the experiments. Note that the algorithms differ only in the execution of line 8: for MBPO the simulated transitions are single $X$-step trajectories starting from real states (i.e. representing sampling new states); and for MBMA the simulated transitions are $X$ single-steps starting from each on-policy state (i.e. representing sampling new actions at visited states). Below we describe the algorithms used in our experiments and discuss their bias-variance structure.

\paragraph{PPO} Proximal Policy Optimization (PPO) \cite{schulman2017proximal} is a model-free on-policy SPG algorithm that leverages multiple actor-critic updates on a single batch of on-policy data. PPO uses a trust-region type of objective that prevents the policy to diverge too much between updates. We use PPO as the single-action agent that performs unbiased policy updates. The algorithm does not reduce the SPG variance beyond using the baseline. As such, we expect PPO variance to be the highest across the tested algorithms.

\begin{algorithm}[hb]
   \caption{MBPO / MBMA with PPO policy}
   \label{alg:algo}
\begin{algorithmic}[1]
   \STATE {\bfseries Input:} batch size $T$, number of simulated samples $X$
   \STATE Collect $T$ on-policy transitions
   \STATE Compute $\lambda$ returns
   \STATE Add $T$ transitions to experience buffer
   \FOR{$i=1$ {\bfseries to} (Epochs * Minibatches)}
   \STATE Update dynamics model on buffer data
   \ENDFOR
   \STATE Simulate $T * X$ new transitions
   \STATE Compute $\lambda$ returns for simulated transitions
   \FOR{$i=1$ {\bfseries to} (Epochs * Minibatches)}
   \STATE Update policy on T*(X+1) transitions
   \STATE Update value on T transitions
   \ENDFOR
\end{algorithmic}
\end{algorithm}

\paragraph{MBMA} PPO that leverages the dynamics model to sample additional actions. The algorithm uses simple MLP transition and reward networks that are trained using MSE loss before performing actor updates. Similarly to QMA, the algorithm performs biased policy updates, with the bias stemming only from the dynamics model Q-value approximation error. Since the dynamics model rollouts depend on the sampled actions, the Q-value approximation has a non-zero variance.

\paragraph{QMA} PPO that uses an auxiliary Q-network to sample additional actions for every visited state\cite{asadi2017mean, petit2019all, ciosek2020expected}. To stabilize the training, we implement QMA-PPO using two Q-networks and choose the smaller prediction for a given state-action pair \cite{van2016deep, haarnoja2018soft}. Q-networks are trained using MSE loss using TD($\lambda$) as targets, which we found to be performing better on average than expected SARSA proposed in the literature \cite{petit2019all, ciosek2020expected}. The updates performed by QMA are biased, as they depend on the output of a biased Q-network. Q-network determinism reduces the absolute variance beyond the reduction stemming from many-actions.

\begin{figure*}[ht]
\begin{center}
\begin{subfigure}{\linewidth}
    \centering
    \includegraphics[width=0.99\linewidth]{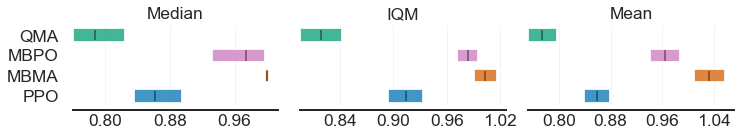}
    \caption{Aggregate performance metrics}\label{fig:image12_reb}
    \vspace{0.15in}
\end{subfigure}
\bigskip
\begin{subfigure}{\linewidth}
    \centering
    \includegraphics[width=0.99\linewidth]{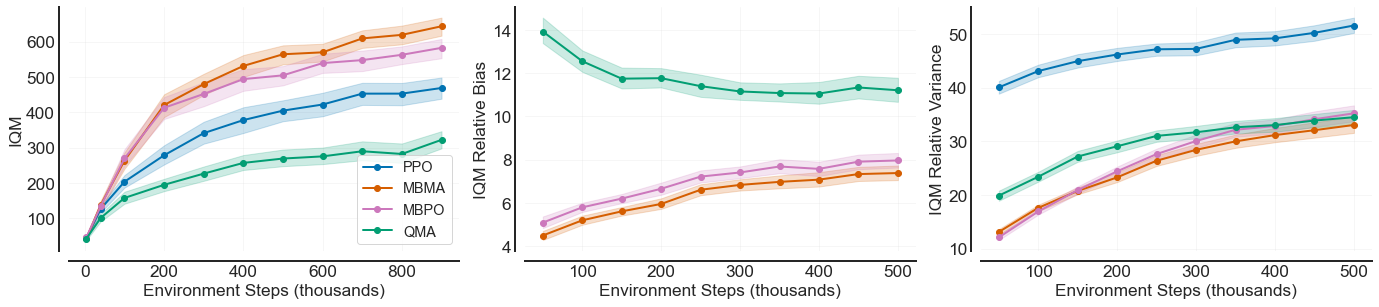}
    \caption{Performance, bias and variance}\label{fig:image12_reb2}
\end{subfigure} 
\vspace{-0.3in}
\caption{Agent performance, bias and variance on DMC-14 (15 seeds, 95\% bootstrapped C.I.). We observe that MBMA generates less bias than other methods for comparable variance reduction effects. Because agents differ only in bias-variance of their policy gradient (\textit{ceteris paribus}), the performance differences stem solely from the beneficial bias-variance structure of the MBMA approach. Furthermore, we observe that the average bias gain of QMA overwhelms its variance reduction translating to worse performance than other algorithms.}
\label{fig1_reb}
\end{center}
\vspace{-0.2in}
\end{figure*}

\paragraph{MBPO} PPO that leverages dynamics model to perform finite horizon rollouts branching from the on-policy data \citep{janner2019trust}. MBPO allows estimating SPG using a mix of real and  simulated states (ie. extend the trajectory length). As such, the algorithm leverages the most common paradigm in model-based SPG - using the dynamics model to generate trajectories \cite{hafner2019dream, kaiser2019model}. Similarly to MBMA, transition and reward networks are trained using MSE loss. Using dynamics model-generated trajectories for SPG updates biases the gradient in two ways. Firstly, similarly to QMA and MBMA, there is bias stemming from Q-value approximation. Secondly, contrary to MA methods, SPG is calculated at states simulated by the model. Due to the extended trajectory, the gradient updates have reduced variance.

We base our implementations on the PPO codebase provided by CleanRL \cite{huang2022cleanrl} and hyperparameters optimized for PPO \citet{shengyi2022the37implementation}. To accommodate more complex tasks, we increase the number of parameters in actor and critic networks across all tasks. Furthermore, we do not use advantage normalization: it has no grounding in SPG theory and can impact the variance structure of the problem at hand; but it can also adversely impact learning in certain environments \cite{andrychowicz2021matters}. All algorithms use the same number of parameters in the actor and critic networks, which are updated the same number of times. QMA, MBPO, and MBMA use an equal number of additional samples (which are tuned for best performance of baselines, see Appendix \ref{appendix21}): for QMA and MBMA we use additional $8$ actions per state; for MBPO we sample rollout of $8$ states per state (which results in extending the trajectory $9$-fold) and $TD(\lambda)$ for value estimation. We anneal the number of additional samples until $15\%$ step of the training for all methods. Whereas learning dynamics models from images is known to work \cite{hafner2019dream, schrittwieser2020mastering}, it is known to offer performance benefits over model-free counterparts stemming from backpropagation of additional non-sparse loss functions \cite{jaderberg2016reinforcement, schwarzer2020data, yarats2021improving}. To mitigate such benefits for algorithms using dynamics models, we use proprioceptive representations given by the environment, with transition and reward networks working on such representations. Similarly, neither MBPO nor MBMA uses an ensemble of dynamics models \citep{buckman2018sample, kurutach2018model, janner2019trust}. Note, that using the same number of simulated samples for all methods yields different computational costs for each method. Calculating Q-value with a dynamics model requires unrolling the model for multiple steps (forward pushes) before bootstrapping it with the critic. In contrast, QMA calculates them in a single step. Relative compute time measurements, hyperparameters, and used netwokrk architectures are detailed in Appendix \ref{appendix3}.

\subsection{Agent Performance, Bias, and Variance}

\begin{table*}[t]
\centering
\caption{IQM PPO normalized performance, bias gain (ie. the amount of bias gained as compared to PPO), and variance reduction (ie. the amount of variance reduced as compared to PPO) of the tested approaches. We bold the best-in-class result. $15$ seeds.}
\vskip 0.1in
 {\renewcommand{\arraystretch}{1.03}%
\begin{tabular}{||l||c|c|c||c|c|c||c|c|c||}
\hline
    & \multicolumn{3}{c||}{\textsc{PPO Normalized Score}} & \multicolumn{3}{c||}{\textsc{Bias gain}} & \multicolumn{3}{c||}{\textsc{Variance reduction}} \\ \hline
    \textsc{task}& \textsc{MBMA} & \textsc{MBPO} & \textsc{QMA} & \textsc{MBMA} & \textsc{MBPO} & \textsc{QMA} & \textsc{MBMA} & \textsc{MBPO} & \textsc{QMA} \\
    \hline\hline
    \textsc{acrobot swingup} & $\mathbf{1.16}$ & $1.11$ & $0.61$ & $\mathbf{8.09}$ & $8.71$ & $10.6$ & $13.0$ & $\mathbf{13.9}$ & $13.2$ \\
    \textsc{ball catch}  & $\mathbf{1.03}$ & $1.02$ & $0.98$ & $\mathbf{5.10}$ & $5.94$ & $12.4$ & $28.7$ & $\mathbf{30.5}$ & $28.9$ \\  
    \textsc{cart swingup} & $\mathbf{1.08}$ & $1.06$ & $1.02$ & $\mathbf{3.23}$ & $3.48$ & $9.04$ & $14.2$ & $\mathbf{14.3}$ & $11.1$ \\
    \textsc{cart 2-poles} & $\mathbf{2.05}$ & $1.33$ & $1.09$ & $\mathbf{6.38}$ & $6.80$ & $10.1$ & $19.0$ & $\mathbf{20.3}$ & $10.3$ \\
    \textsc{cart 3-poles} & $0.98$ & $\mathbf{1.26}$ & $1.15$ & $7.88$ & $\mathbf{7.67}$ & $11.0$ & $\mathbf{15.4}$ & $13.1$ & $10.9$ \\
    \textsc{cheetah run} & $\mathbf{1.82}$ & $1.74$ & $0.77$ & $\mathbf{11.4}$ & $12.0$ & $21.3$ & $38.6$ & $\mathbf{39.5}$ & $26.0$ \\
    \textsc{finger spin} & $0.87$ & $0.79$ & $\mathbf{0.88}$ & $\mathbf{4.49}$ & $4.88$ & $9.58$ & $\mathbf{14.0}$ & $3.81$ & $10.5$ \\
    \textsc{finger turn} & $\mathbf{1.02}$ & $0.99$ & $0.88$ & $\mathbf{3.45}$ & $4.05$ & $10.1$ & $\mathbf{23.0}$ & $18.1$ & $20.2$ \\
    \textsc{point easy} & $\mathbf{1.01}$ & $1.00$ & $0.76$ & $\mathbf{1.36}$ & $1.50$ & $3.91$ & $11.2$ & $\mathbf{11.9}$ & $11.3$ \\
    \textsc{reacher easy} & $1.03$ & $\mathbf{1.04}$ & $0.68$ & $\mathbf{3.94}$ & $4.64$ & $10.2$ & $22.3$ & $\mathbf{22.8}$ & $21.7$ \\
    \textsc{reacher hard} & $1.39$ & $\mathbf{1.40}$ & $0.75$ & $\mathbf{5.29}$ & $6.20$ & $11.7$ & $20.7$ & $\mathbf{21.7}$ & $18.6$ \\
    \textsc{walker stand} & $\mathbf{1.03}$ & $1.02$ & $0.96$ & $\mathbf{9.70}$ & $11.5$ & $19.2$ & $\mathbf{29.8}$ & $26.6$ & $18.9$ \\
    \textsc{walker walk} & $\mathbf{1.76}$ & $1.46$ & $1.01$ & $\mathbf{11.5}$ & $12.7$ & $16.2$ & $\mathbf{37.2}$ & $35.5$ & $17.5$ \\
    \textsc{walker run} & $\mathbf{1.67}$ & $1.19$ & $1.05$ & $\mathbf{11.3}$ & $12.2$ & $15.8$ & $36.7$ & $\mathbf{37.5}$ & $17.5$ \\
    \hline\hline
\end{tabular}}
\label{maasdc_grid00}
\vskip -0.1in
\end{table*}

We compare the performance of agents on $14$ DMC tasks \cite{tassa2018deepmind} of varying difficulty for $1M$ environments steps and $15$ seeds. During this training, we measure agent performance, as well as bias and variance of policy gradients. Furthermore, to measure algorithms performance in longer training regimes, we record agent performance on $4$ difficult DMC tasks (quadruped walk; quadruped run; humanoid stand; and humanoid walk) for $3M$ and $6M$ environment steps respectively. We record robust statistics \cite{agarwal2021deep} for all runs. We provide detailed results, methodology for calculating bias and variance, and further experimental details in Appendix \ref{appendix3}. 

We find that MBMA performs better in $14$ out of $18$ DMC tasks, while MBPO and PPO have better performance in $3$ and $1$ environments respectively. However, the performance differences are within the margin of statistical error for some cases. Note that we use hyperparameters tuned wrt. PPO and MBPO. We observe greater performance gaps benefiting MBMA for different hyperparameter settings (see Appendix \ref{appendix21}, where we compare the performance for different numbers of simulated samples and various simulation horizons).

\begin{table}[h]
\caption{IQM on four complex DMC tasks (8 seeds, 1 std of the mean). 3$M$ and 6$M$ steps for quadruped and humanoid tasks respectively.}
\vskip 0.1in
\label{sample-table}
\begin{center}
{\renewcommand{\arraystretch}{1.2}%
\begin{tabular}{||l||c|c|c||}
\hline
 & \textsc{PPO} & \textsc{MBMA} & \textsc{MBPO}  \\
\hline\hline
\textsc{quad walk} & $667\pm31$ & $\mathbf{677\pm30}$ & $590\pm43$ \\
\textsc{quad run} & $455\pm18$ & $\mathbf{468\pm6}$ & $460\pm20$ \\
\textsc{hum stand} & $189\pm8$ & $\mathbf{214\pm2}$ & $203\pm31$ \\
\textsc{hum walk} & $178\pm14$ & $\mathbf{210\pm8}$ & $198\pm16$\\
\hline
\end{tabular}}
\end{center}
\vskip -0.1in
\end{table}

In line with theory, we find that MBMA produces consistently less bias than other methods while offering greater or comparable variance reduction. On average, MBMA measures the lowest bias and lowest variance. Furthermore, we find that QMA produces smaller gradients than other methods given the same data. This points towards the low variation of the Q-network output and subsequent gradient cancellation. We find that QMA has the highest relative bias despite the MA approach. We find this unsurprising, since as noted in earlier sections, Q-networks pursue a more difficult target than dynamics models. Furthermore, even though QMA has the lowest absolute variance (due to no stochasticity in Q-value estimation), its smallest expected gradient size leads to a greater impact on its variance and thus has the highest relative variance amongst methods.

\section{Related Work}

\subsection{Many-Actions SPG} 

The idea of sampling many actions per state was proposed in an unfinished preprint\footnote{http://incompleteideas.net/papers/SSM-unpublished.pdf} by Sutton et al. (2001). Later, the topic was expanded upon by several authors. TRPO \cite{schulman2015trust} 'vine procedure' uses multiple without-replacement action samples per state generated via environment rewinding. The without-replacement PG estimator was further refined by using the without-replacement samples as a free baseline \citep{kool2019estimating, kool2019buy}. MAC \cite{asadi2017mean} calculates the inner integral of SPG exactly (ie. sample the entire action space for given states) using Q-network, with the scheme applicable only to discrete action spaces and tested on simple environments. Similarly, \citet{petit2019all} propose to estimate the inner integral with a quadrature of $N$ samples given by a Q-network. The authors also derive the basic theoretical properties of MA SPG. Besides expanding on the theoretical framework, \citet{ciosek2020expected} propose an off-policy algorithm that, given a Gaussian actor and quadratic critic, can compute the inner integral analytically. 

\subsection{Model-Based RL}

ME-TRPO \cite{kurutach2018model} leverages an ensemble of environment models to increase the sample efficiency of TRPO. WM \cite{ha2018recurrent} uses environment interactions to learn the dynamics model, with the policy learning done via evolutionary strategies inside the dynamics model. Similarly, SimPLe  \cite{kaiser2019model} learns the policy by simulating states via the dynamics model. Dreamer \cite{hafner2019dream, hafner2020mastering} refines the latent dynamics model learning by proposing a sophisticated joint learning scheme for recurrent transition and discrete state representation models, but the policy learning is still done by simulating states inside the dynamics model starting from sampled off-policy transitions. Notably, Dreamer was shown to solve notoriously hard Humanoid task \cite{yarats2021mastering}. Differentiable dynamics models allow for direct gradient optimization of the policy as an alternative to traditional SPG. Methods like MAAC \cite{clavera2019model} and DDPPO \cite{li2022gradient} explore policy optimization via backpropagating through the dynamics model. MuZero \cite{schrittwieser2020mastering} leverages the dynamics model to perform a Monte-Carlo tree search inside the latent model. Perhaps the closest to the proposed approach is MBVE \cite{feinberg2018model}. There, an off-policy DPG agent uses the dynamics model to estimate $n$-step Q-values and thus refine the approximation. However, our analysis is restricted to model-based on-policy SPG and we leave the analysis of MBMA in the context of off-policy agents and backpropagating through dynamics model for future work.

\section{Conclusions}

In this paper, we analyzed the variance of the SPG estimator mathematically. We showed that it can be disaggregated into sub-components dependent on policy stochasticity, as well as the components which are dependent solely on the structure of the Markov process underlying the policy-embedded MDP. By optimizing such components with respect to the number of state and action samples, we derived an optimality condition that shows when MA is a preferable strategy as compared to traditional, single-action SPG. We used the result to show the difficult conditions MA has to meet to be an optimal choice for the case of contextual bandit problems. We hope that those theoretical results will reinvigorate research into MA estimation in the context of RL.

Furthermore, we discussed the bias-variance trade-off induced by using Q-network and dynamics models to simulate action or state samples. We showed that the bias associated with simulating additional states is of more complex form than the bias associated with simulating actions while offering similar variance reduction benefits. We measured the relative bias and variance of policy gradients calculated via each method and found the measurements in line with theoretical predictions, showing the analytical importance of bias and variance of SPG. We hope that those results will impact the domain of model-based on-policy SPG, where leveraging the dynamics model for trajectory simulation is the dominating approach for stochastic policy gradient. 

Finally, we proposed an MBMA module - an approach that leverages dynamics models for MA estimation at the cost of additional computations. We evaluated its performance against QMA, MBPO, and PPO on-policy baselines. Our experiments showed that it compares favorably in terms of both sample efficiency and final performance in most of the tested environments. We release the code used for experiments under the following address \href{https://github.com/naumix/On-Many-Actions-Policy-Gradient}{https://github.com/naumix/On-Many-Actions-Policy-Gradient}.

\section{Limitations}

The main limitation of our theoretical analysis is its dependence on the Markov chain Central Limit Theorem, as such its results hold only if the underlying Markov chain is ergodic. Furthermore, it is conducted in the context of on-policy SPG and its conclusions are applicable only to such settings. Following the theoretical analysis, our experiments tested only on-policy SPG algorithms. We consider expanding MA analysis to off-policy setting an interesting avenue for future research.

\section{Acknowledgements}

We would like to thank Witold Bednorz, Piotr Miłoś, and Łukasz Kuciński for valuable discussions and notes. Marek Cygan is cofinanced by National Centre for Research and Development as a part of EU supported Smart Growth Operational Programme 2014-2020 (POIR.01.01.01-00-0392/17-00). The experiments were performed using the Entropy cluster funded by NVIDIA, Intel, the Polish National Science Center grant UMO-2017/26/E/ST6/00622, and ERC Starting Grant TOTAL.

\bibliography{icml2023/example_paper}
\bibliographystyle{icml2023}

\newpage
\appendix
\onecolumn

\section{Derivations - Variance}
\label{appendix1}

First, we augment the notation to encompass many action samples:

\begin{equation*}
    \Upsilon^{t}_{s,a^{n}} = \nabla J(s_t, a_{t}^{n}), \quad \Upsilon^{t}_{s,a} = \nabla J(s_t, a_{t}) \quad \text{and} \quad \Upsilon^{t}_{s} = \underset{a \sim \pi}{\mathbb{E}} ~ \Upsilon^{t}_{s,a}
\end{equation*} 

For convenience, throughout the Appendix we will assume finite state and action spaces. However, the same reasoning works for continuous spaces.

\subsection{Derivation of Lemma \ref{lem:lemma1}}
\label{appendix11}

Following the MA-SPG definition outlined in Equation \ref{eq:sum}, $\mathrm{Var}_{s, a \sim p^{\pi}_{0}, \pi} ~ \bigl[\Upsilon_{s,a}\bigr]$ is equal to:

\begin{equation}
\label{eq:var1}
\begin{split}
    \underset{s, a \sim p^{\pi}_{0}, \pi}{\mathrm{Var}} ~ \bigl[\Upsilon_{s,a}\bigr] & = \sum_{s} ~ p^{\pi}_{0}(s) ~ \prod_{n=1}^{N} ~ \sum_{a^n} \pi(a^n|s) ~ \biggl( \frac{\Upsilon_{s,a^1}}{N} + ... + \frac{\Upsilon_{s,a^N}}{N} \biggr)^2 - \biggl( \mathbb{E} ~ \nabla J \biggr)^2\\
    & = \frac{N}{N^2} ~ \sum_{s} ~ p^{\pi}_{0}(s) ~ \sum_{a} \pi(a|s) ~ (\Upsilon_{s,a})^2 + \frac{2}{N^2} ~ \binom{N}{2} ~ \sum_{s} ~ p^{\pi}_{0}(s) ~ \biggl( \sum_a ~ \pi(a|s) ~ \Upsilon_{s,a} \biggr)^2 - \biggl( \mathbb{E} ~ \nabla J \biggr)^2 \\
    & = \frac{1}{N} ~ \underset{s \sim p^{\pi}_{0}}{\mathbb{E}} ~ \underset{a \sim \pi}{\mathbb{E}} ~ (\Upsilon_{s,a})^2 + \frac{N-1}{N} ~ \underset{s \sim p^{\pi}_{0}}{\mathbb{E}} ~ (\Upsilon_{s})^2 - \biggl( \mathbb{E} ~ \nabla J \biggr)^2 \\
    & = \frac{1}{N} ~ \underset{s \sim p^{\pi}_{0}}{\mathbb{E}} ~ \underset{a \sim \pi}{\mathbb{E}} ~ (\Upsilon_{s,a})^2 + \frac{N-1}{N} ~ \underset{s \sim p^{\pi}_{0}}{\mathbb{E}} ~ (\Upsilon_{s})^2 - \biggl( \mathbb{E} ~ \nabla J \biggr)^2 \\
    & = \frac{1}{N} ~ \underset{s \sim p^{\pi}_{0}}{\mathbb{E}} ~ \underset{a \sim \pi}{\mathbb{E}} ~ (\Upsilon_{s,a})^2 + \frac{N-1}{N} ~ \underset{s \sim p^{\pi}_{0}}{\mathbb{E}} ~ (\Upsilon_{s})^2 - \biggl( \mathbb{E} ~ \nabla J \biggr)^2 \\
    & = \frac{1}{N} \biggl( \underset{s \sim p^{\pi}_{0}}{\mathbb{E}} ~ \underset{a \sim \pi}{\mathbb{E}} ~ (\Upsilon_{s,a})^2 - \underset{s \sim p^{\pi}_{0}}{\mathbb{E}} ~ (\Upsilon_{s})^2  \biggr) + \underset{s \sim p^{\pi}_{0}}{\mathbb{E}} ~ (\Upsilon_{s})^2 - \biggl( \mathbb{E} ~ \nabla J \biggr)^2 \\
    & = \underset{s \sim p^{\pi}_{0}}{\mathrm{Var}} ~ \bigl[ \Upsilon_s \bigr] + \frac{1}{N} \underset{s \sim p^{\pi}_{0}}{\mathbb{E}} ~ \underset{a \sim \pi}{\mathrm{Var}} ~ \bigl[ \Upsilon_{s,a} \bigr] \\
    & = \underset{s_0 \sim p^{\pi}_{0}}{\mathrm{Var}} ~ \bigl[ \underset{a_0 \sim \pi}{\mathbb{E}} \nabla_{\theta}J(s_0,a_0) \bigr] + \frac{1}{N} \underset{s_0 \sim p^{\pi}_{0}}{\mathbb{E}} ~ \underset{a_0 \sim \pi}{\mathrm{Var}} ~ \bigl[ \nabla J(s_0,a_0) \bigr]
\end{split}
\end{equation}

The above result for $N=1$ is reported in \cite{petit2019all}, noting it as stemming from the law of total variance. However, we could not find the proof in the existing literature. Below, $p^{\pi}_{t}(s_t|s_0, a_{0}^{1})$ denotes the $t$ step transition kernel conditioned on $s_0$ and $a_{0}^{1}$ (ie. the first sampled action in $s_0$). 

\begin{equation*}
\begin{split}
    & \mathbb{E} \bigl[ \Upsilon_{s,a} \Upsilon_{s,a}^{t} \bigr] = \\
    & = \sum_{s_0} p^{\pi}_{0}(s_0) \prod_{n=1}^{N} \sum_{a^{n}_{0}} \pi(a^{n}_{0}|s_0) \sum_{s_t} p^{\pi}_{t}(s_t|s_0, a_{0}^{1}) \prod_{m=1}^{N} \sum_{a^{m}_{t}} \pi(a^{m}_{t}|s_t) \biggl( \frac{\Upsilon_{s,a^1}}{N} + ... + \frac{\Upsilon_{s,a^N}}{N} \biggr) ~ \biggl( \frac{\Upsilon_{s,a^1}^{t}}{N} + ... + \frac{\Upsilon_{s,a^N}^{t}}{N} \biggr) \\
    & = \sum_{s_0} ~ p^{\pi}_{0}(s_0) ~ \prod_{n=1}^{N} ~ \sum_{a^{n}_{0}} \pi(a^{n}_{0}|s_0) ~ \sum_{s_t} p^{\pi}_{t}(s_t|s_0,a_{0}^{1}) \prod_{m=1}^{N} ~ \sum_{a^{m}_{t}} \pi(a^{m}_{t}|s_t) \biggl( \sum_{i = 1}^{N} ~ \sum_{j = 1}^{N} ~ \frac{\Upsilon_{s,a^{i}}}{N} \frac{\Upsilon_{s,a^{j}}^{t}}{N} \biggr) \\
\end{split}
\end{equation*}

Therefore:

\begin{equation*}
\begin{split}    
    \mathbb{E} \bigl[ \Upsilon_{s,a} \Upsilon_{s,a}^{t} \bigr] = & = \frac{1}{N} ~ \sum_{s_0} ~ p^{\pi}_{0}(s_0) ~ \sum_{a_{0}^{1}} ~ \pi(a_{0}^{1}) ~ \Upsilon_{s,a_{0}^{1}} ~ \prod_{n=2}^{N} ~ \sum_{a^{n}_{0}} \pi(a^{n}_{0}|s_0) ~ \sum_{s_t} p^{\pi}_{t}(s_t|s_0, a_{0}^{1}) ~ \sum_{a_{t}} \pi(a_{t}|s_t) ~ \Upsilon_{s,a}^{t} \\
    & + \frac{N-1}{N} ~ \sum_{s_0} ~ p^{\pi}_{0}(s_0) ~ \sum_{a_{0}^{2}} ~ \pi(a_{0}^{2}) ~ \Upsilon_{s,a_{0}^{2}} ~ \sum_{a^{1}_{0}} \pi(a^{1}_{0}|s_0) ~ \sum_{s_t} p^{\pi}_{t}(s_t|s_0,a_{0}^{1}) ~ \sum_{a_{t}} \pi(a_{t}|s_t) ~ \Upsilon_{s,a}^{t} \\
    & = \frac{1}{N} ~ \sum_{s_0} ~ p^{\pi}_{0}(s_0) ~ \sum_{a_{0}^{1}} ~ \pi(a_{0}^{1}) ~ \Upsilon_{s,a_{0}^{1}} ~ \sum_{s_t} ~ p^{\pi}_{t}(s_t|s_0, a_{0}^{1}) ~ \Upsilon_{s}^{t} \\
    & + \frac{N-1}{N} ~ \sum_{s_0} ~ p^{\pi}_{0}(s_0) ~ \Upsilon_{s} ~ \sum_{a^{1}_{0}} ~ \pi(a^{1}_{0}|s_0) ~ \sum_{s_t} ~ p^{\pi}_{t}(s_t|s_0, a_{0}^{1}) ~ \Upsilon_{s}^{t}
\end{split}
\end{equation*}

Thus, the $t^{th}$ covariance of MA is equal to:

\begin{equation}
\label{eq:cov1}
\begin{split}
    & \underset{s_t, a_t \sim p^{\pi}_{t}, \pi}{\mathrm{Cov}} ~ \left[ \Upsilon_{s,a}, \Upsilon_{s,a}^{t} \right] = \\
    & = \frac{1}{N} ~ \sum_{s_0} ~ p^{\pi}_{0}(s_0) ~ \sum_{a_{0}} ~ \pi(a_{0}) ~ \Upsilon_{s,a} ~ \sum_{s_t} ~ p^{\pi}_{t}(s_t|s_0, a_{0}) ~ \Upsilon_{s}^{t} \\
    & + \frac{N-1}{N} ~ \sum_{s_0} ~ p^{\pi}_{0}(s_0) ~ \Upsilon_{s} ~ \sum_{a_{0}} ~ \pi(a_{0}|s_0) ~ \sum_{s_t} ~ p^{\pi}_{t}(s_t|s_0, a_{0}) ~ \Upsilon_{s}^{t} \\
    & - \biggl( \sum_{s_0} ~ p^{\pi}_{0}(s_0) ~ \sum_{a_{0}} ~ \pi(a_{0}) ~ \Upsilon_{s,a} \biggr) \biggl( \sum_{s_t} ~ p^{\pi}_{t}(s_t) ~ \sum_{a_{t}} ~ \pi(a_{t}) ~ \Upsilon_{s,a}^{t} \biggr) \\
    & = \frac{1}{N} ~ \sum_{s_0} ~ p^{\pi}_{0}(s_0) ~ \sum_{a_{0}} ~ \pi(a_{0}) ~ \Upsilon_{s,a} ~ \sum_{s_t} ~ p^{\pi}_{t}(s_t|s_0, a_{0}) ~ \Upsilon_{s}^{t} \\
    & + \frac{N-1}{N} ~ \sum_{s_0} ~ p^{\pi}_{0}(s_0) ~ \Upsilon_{s} ~ \sum_{a_{0}} ~ \pi(a_{0}|s_0) ~ \sum_{s_t} ~ p^{\pi}_{t}(s_t|s_0, a_{0}) ~ \Upsilon_{s}^{t} - \bigl( \underset{a \sim \pi}{\mathbb{E}} \Upsilon_{s,a} \bigr) \bigl( \underset{a \sim \pi}{\mathbb{E}}  \Upsilon_{s,a}^{t} \bigr) \\
    & = \frac{1}{N} \underset{s_0 \sim p^{\pi}_{0}}{\mathbb{E}} \biggl( \sum_{a_{0}} ~ \pi(a_{0}) ~ \Upsilon_{s,a}^{0} ~ \sum_{s_t} ~ p^{\pi}_{t}(s_t|s_0, a_{0}) ~ \Upsilon_{s}^{t} - \Upsilon_{s}^{0} ~ \sum_{a_{0}} ~ \pi(a_{0}|s_0) ~ \sum_{s_t} ~ p^{\pi}_{t}(s_t|s_0, a_{0}) ~ \Upsilon_{s}^{t} \biggr) \\
    & + \biggl( \sum_{s_0} ~ p^{\pi}_{0}(s_0) ~ \Upsilon_{s}^{0} ~ \sum_{a_{0}} ~ \pi(a_{0}|s_0) ~ \sum_{s_t} ~ p^{\pi}_{t}(s_t|s_0, a_{0}) ~ \Upsilon_{s}^{t} - \bigl( \underset{a \sim \pi}{\mathbb{E}} \Upsilon_{s,a} \bigr) \bigl( \underset{a \sim \pi}{\mathbb{E}} \Upsilon_{s,a}^{t} \bigr) \biggr) \\
    & = \underset{s_t, a_t \sim p_{t}^{\pi}, \pi}{\mathrm{Cov}} ~ \bigl[ \Upsilon_{s}, \Upsilon_{s}^{t} \bigr] +  \frac{1}{N} \underset{s_0 \sim p^{\pi}_{0}}{\mathbb{E}} ~ \underset{s_t, a_t \sim p_{t}^{\pi}, \pi}{\mathrm{Cov}} \bigl[ \Upsilon_{s,a}, \Upsilon_{s,a}^{t}\bigr] \\
    & = \underset{s_t, a_t \sim p_{t}^{\pi}, \pi}{\mathrm{Cov}} ~ \bigl[ \underset{a_0 \sim \pi}{\mathbb{E}} \nabla_{\theta}J(s_0,a_0), \underset{a_0 \sim \pi}{\mathbb{E}} \nabla_{\theta}J(s_t,a_t) \bigr] +  \frac{1}{N} \underset{s_0 \sim p^{\pi}_{0}}{\mathbb{E}} ~ \underset{s_t, a_t \sim p_{t}^{\pi}, \pi}{\mathrm{Cov}} \bigl[ \nabla J(s_0, a_0), \nabla J(s_t, a_{t})\bigr]
\end{split}
\end{equation}

Combining Equations \ref{eq:var1} and \ref{eq:cov1} concludes derivation of Lemma \ref{lem:lemma1}. 

\subsection{Derivation of Lemma \ref{lemma2}}
\label{appendix12}
Since $N$ is defined to be a natural number, we calculate the variance reduction effect stemming from increasing $N$ via the forward difference operator:

\begin{equation*}
\begin{split}
    \Delta_N = \mathbf{V}(N+1) - \mathbf{V}(N)
\end{split}
\end{equation*}

We also use the shorthand notation:

\begin{equation*}
\begin{split}
    \alpha^{t}_{e} = \underset{s_t, a_t \sim p^{\pi}_{t}, \pi}{\mathrm{Cov}} \bigl[\Upsilon_{s}, \Upsilon_{s}^{t}\bigr], \quad \alpha^t & = \underset{s_t, a_t \sim p_{t}^{\pi}, \pi | s_0}{ \mathrm{Cov}} \bigl[ \Upsilon_{s,a}, \Upsilon_{s,a}^{t}\bigr] \quad \text{and} \quad
    \mathrm{C}^t = \underset{s_t, a_t \sim p^{\pi}_{t}, \pi}{\mathrm{Cov}} ~ \bigl[ \Upsilon_{s,a}, \Upsilon_{s,a}^{t} \bigr] = \alpha^{t}_{e} + \frac{1}{N} ~ \underset{s \sim p^{\pi}_{0}}{\mathbb{E}} ~ \alpha^{t}
\end{split}
\end{equation*}

Thus:

\begin{equation*}
\label{eq:varrrr2}
\begin{split}
    & \mathbf{V} = \frac{1}{T} \biggl( \underset{s_0 \sim p^{\pi}_{0}}{\mathrm{Var}} ~ \bigl[\Upsilon_{s}\bigr] +  2\sum_{t=1}^{T-1} ~ \frac{T-t}{T} ~ \alpha^{t}_{e}  + \frac{1}{N} \underset{s_0 \sim p^{\pi}_{0}}{\mathbb{E}} \bigl( ~ \underset{a \sim \pi}{\mathrm{Var}} ~ \bigl[\Upsilon_{s,a}^{0}\bigr]
    + 2\sum_{t=1}^{T-1} ~ \frac{T-t}{T} ~ \alpha^{t} \bigr) \biggr)
\end{split}
\end{equation*}

We proceed with the calculation of the forward difference:

\begin{equation}
\label{eq:lem2}
\begin{split}
    \Delta_N & = \frac{1}{T} \biggl( \underset{s \sim p^{\pi}_{0}}{\mathrm{Var}} ~ \bigl[\Upsilon_{s}\bigr] +  2\sum_{t=1}^{T-1} ~ \frac{T-t}{T} ~ \alpha^{t}_{e} + \frac{1}{N+1} \underset{s \sim p^{\pi}_{0}}{\mathbb{E}} \bigl( ~ \underset{a \sim \pi}{\mathrm{Var}} ~ \bigl[\Upsilon_{s,a}\bigr]
    + 2\sum_{t=1}^{T-1} ~ \frac{T-t}{T} ~ \alpha^{t} \bigr) \biggr) \\
    & - \frac{1}{T} \biggl( \underset{s \sim p^{\pi}_{0}}{\mathrm{Var}} ~ \bigl[\Upsilon_{s}\bigr] +  2\sum_{t=1}^{T-1} ~ \frac{T-t}{T} ~ \alpha^{t}_{e} + \frac{1}{N} \underset{s \sim p^{\pi}_{0}}{\mathbb{E}} \bigl( ~ \underset{a \sim \pi}{\mathrm{Var}} ~ [\Upsilon_{s,a}]
    + 2\sum_{t=1}^{T-1} ~ \frac{T-t}{T} ~ \alpha^{t} \bigr) \biggr) \\
    & = \frac{1}{T(N+1)} \underset{s \sim p^{\pi}_{0}}{\mathbb{E}} \biggl( ~ \underset{a \sim \pi}{\mathrm{Var}} ~ \bigl[\Upsilon_{s,a}\bigr]
    + 2\sum_{t=1}^{T-1} ~ \frac{T-t}{T} ~ \alpha^{t} \biggr) \\
    & - \frac{1}{T~N} \underset{s \sim p^{\pi}_{0}}{\mathbb{E}} \biggl( ~ \underset{a \sim \pi}{\mathrm{Var}} ~ \bigl[\Upsilon_{s,a}\bigr]
    + 2\sum_{t=1}^{T-1} ~ \frac{T-t}{T} ~ \alpha^{t} \biggr) \\
    & = \frac{-1}{T(N^2+N)} \underset{s \sim p^{\pi}_{0}}{\mathbb{E}} \biggl( ~ \underset{a \sim \pi}{\mathrm{Var}} ~ \bigl[\Upsilon_{s,a}\bigr]
    + 2\sum_{t=1}^{T-1} ~ \frac{T-t}{T} ~ \alpha^{t} \biggr) \\
    & = \frac{-1}{T(N^2+N)} \underset{s \sim p^{\pi}_{0}}{\mathbb{E}} \biggl( ~ \underset{a \sim \pi}{\mathrm{Var}} ~ \bigl[\Upsilon_{s,a}\bigr]
    + 2\sum_{t=1}^{T-1} ~ \frac{T-t}{T} ~ \underset{s_t, a_t \sim p_{t}^{\pi}, \pi | s_0}{ \mathrm{Cov}} \bigl[ \Upsilon_{s,a}^{0}, \Upsilon_{s,a}^{t}\bigr]  \biggr) \\
    & = \frac{-1}{T(N^2+N)} \underset{s_0 \sim p^{\pi}_{0}}{\mathbb{E}} \biggl( ~ \underset{a \sim \pi}{\mathrm{Var}} ~ \bigl[\nabla J(s_0, a_0)\bigr]
    + 2\sum_{t=1}^{T-1} ~ \frac{T-t}{T} ~ \underset{s_t, a_t \sim p_{t}^{\pi}, \pi | s_0}{ \mathrm{Cov}} \bigl[ \nabla J(s_0, a_0), \nabla J(s_t, a_t)\bigr]  \biggr)
\end{split}
\end{equation}

Similarly, we calculate $\Delta_T$:

\begin{equation*}
\label{eq:vt}
\begin{split}
    \Delta_T & = \frac{1}{T+ \delta T} \underset{s, a \sim p^{\pi}_{0}, \pi}{\mathrm{Var}} [\Upsilon_{s,a}] + 2 \sum_{t=1}^{T + \delta T -1} \frac{T+\delta T-t}{(T+\delta T)^2} ~ \mathrm{C}^{t} - \frac{1}{T} \underset{s, a \sim p^{\pi}_{0}, \pi}{\mathrm{Var}} \bigl[\Upsilon_{s,a}\bigr] - 2 \sum_{t=1}^{T-1} \frac{T-t}{T^2} ~ \mathrm{C}^{t} \\
    & = \frac{-\delta T}{T + \delta T} \underset{s, a \sim p^{\pi}_{0}, \pi}{\mathrm{Var}} \bigl[\Upsilon_{s,a}\bigr] + 2 \sum_{t=1}^{T -1} \biggl( \frac{T+\delta T-t}{(T+\delta T)^2} - \frac{T-t}{T^2} \biggr) ~ \mathrm{C}^{t} +  2 \sum_{k=T}^{T +\delta T - 1} \frac{T-t}{(T+\delta T)^2} ~ \mathrm{C}^{t} \\
    & = \frac{-\delta}{T + \delta T} \biggl( \underset{s, a \sim p^{\pi}_{0}, \pi}{\mathrm{Var}} [\Upsilon_{s,a}] + 2\sum_{t=1}^{T-1} \bigl( \frac{T-t}{T} - \frac{t}{T+\delta T} \bigr) ~ \mathrm{C}^t  - \frac{2}{\delta} \sum_{k=T}^{T + \delta T - 1} \frac{T + \delta T - k}{T + \delta T} ~ \mathrm{C}^k \Biggr)
\end{split}
\end{equation*}

Now, we assume that the trajectory length guarantees reaching a regenerative state, and thus $\sum_{k=T}^{T + \delta T - 1} \frac{T + \delta T - k}{T + \delta T} ~ \mathrm{C}^k = 0$ :

\begin{equation}
\label{eq:vt2}
\begin{split}
    \Delta_T & = \frac{-\delta}{T + \delta T} \biggl( \underset{s, a \sim p^{\pi}_{0}, \pi}{\mathrm{Var}} [\Upsilon_{s,a}] + 2\sum_{t=1}^{T-1} \bigl( \frac{T-t}{T} - \frac{t}{T+\delta T} \bigr) ~ \mathrm{C}^t \Biggr) \\
    & = \frac{-\delta}{T + \delta T} \biggl( \underset{s, a \sim p^{\pi}_{0}, \pi}{\mathrm{Var}} [\Upsilon_{s,a}] + 2\sum_{t=1}^{T-1} \bigl( \frac{T-t}{T} - \frac{t}{T+\delta T} \bigr) ~ \underset{s_t, a_t \sim p^{\pi}_{t}, \pi}{\mathrm{Cov}} ~ \bigl[ \Upsilon_{s,a}^{0}, \Upsilon_{s,a}^{t} \bigr] \Biggr)
\end{split}
\end{equation}

Combining Equations \ref{eq:lem2} and \ref{eq:vt2} concludes derivation of Lemma \ref{lemma2}.

\subsection{Derivation of Theorem 3.3}
\label{appendix13}
We start the derivation by stating that MA-SPG is advantageous in terms of variance reduction as compared to increased trajectory length SPG when $- \Delta_N \geq -\Delta_T$. As such:

\begin{equation*}
\label{eq:t0}
\begin{split}
    \frac{1+ \delta}{\delta (N^2+N)} \underset{s \sim p^{\pi}_{0}}{\mathbb{E}} \biggl( \underset{a \sim \pi}{\mathrm{Var}} ~ \bigl[\Upsilon_{s,a}\bigr]
    + 2\sum_{t=1}^{T-1} ~ \frac{T-t}{T} ~ \alpha^{t} \biggr) \geq \underset{s, a \sim p^{\pi}_{0}, \pi}{\mathrm{Var}} \bigl[\Upsilon_{s,a}\bigr] + 2\sum_{t=1}^{T-1} \bigl( \frac{T-t}{T} - \frac{t}{T+\delta T} \bigr) ~ \mathrm{C}^t
\end{split}
\end{equation*}

We use Equations \ref{eq:var1} and \ref{eq:cov1} to expand the RHS:

\begin{equation*}
\label{eq:t1}
\begin{split}
    & \underset{s, a \sim p^{\pi}_{0}, \pi}{\mathrm{Var}} \bigl[\Upsilon_{s,a}\bigr] + 2\sum_{t=1}^{T-1} \bigl( \frac{T-t}{T} - \frac{t}{T+\delta T} \bigr) ~ \mathrm{C}^t = \\
    & = \underset{s \sim p^{\pi}_{0}}{\mathrm{Var}} ~ [\Upsilon_s] + \frac{1}{N} \underset{s \sim p^{\pi}_{0}}{\mathbb{E}} ~ \underset{a \sim \pi}{\mathrm{Var}} ~ \bigl[\Upsilon_{s,a}\bigr] + 2\sum_{t=1}^{T-1} \bigl( \frac{T-t}{T} - \frac{t}{T+\delta T} \bigr) ~ \bigl(\alpha^{t}_{e} + \frac{1}{N} ~ \alpha^{t} \bigr)
\end{split}
\end{equation*}

We move all terms dependent on the policy to the LHS:

\begin{equation}
\label{eq:theoremsupp}
\begin{split}
    & \frac{1 - \delta N}{\delta (N^2 + N)} \underset{s \sim p^{\pi}_{0}}{\mathbb{E}} \underset{a \sim \pi}{\mathrm{Var}} ~ \bigl[\Upsilon_{s,a}\bigr] + 2\sum_{t=1}^{T-1} \bigl(\frac{(1 + \delta - \delta N - \delta^{2} N) T - (1 - 2 \delta N - \delta^2 N) t}{(\delta T + \delta^2 T)(N^2 + N)} \bigr) ~ \alpha^{t} \geq \\
    & \underset{s \sim p^{\pi}_{0}}{\mathrm{Var}} ~ [\Upsilon_s] + 2\sum_{t=1}^{T-1} \bigl( \frac{T-t}{T} - \frac{t}{T+\delta T} \bigr) ~ \alpha^{t}_{e}
\end{split}
\end{equation}

Now, in order to recover the Corollary \ref{cor1}, we assume a contextual bandit setup (ie. $p^{\pi}(s'|s) = p^{\pi}(s')$). Then:

\begin{equation*}
\begin{split}
    \frac{1 - \delta N}{\delta (N^2 + N)} \underset{s \sim p^{\pi}_{0}}{\mathbb{E}} \underset{a \sim \pi}{\mathrm{Var}} ~ [\Upsilon_{s,a}] \geq \underset{s \sim p^{\pi}_{0}}{\mathrm{Var}} ~ [\Upsilon_s]
\end{split}
\end{equation*}

Which is equivalent to:

\begin{equation*}
    \frac{\underset{s \sim p^{\pi}_{0}}{\mathrm{Var}} ~ [\Upsilon_s]}{\underset{s \sim p^{\pi}_{0}}{\mathbb{E}} \underset{a \sim \pi}{\mathrm{Var}} ~ [\Upsilon_{s,a}]} \leq \frac{1 - \delta N}{\delta (N^2 + N)}
\end{equation*}

We proceed with the derivation for the MDP setup, where $p^{\pi}(s'|s) \neq p^{\pi}(s')$. We write $N = 1$, which implies that we start in the regular single-action SPG setup. Furthermore, we assume $\delta = 1$, which according to the setup implies equal cost of sampling additional action and state samples. Thus, Equation \ref{eq:theoremsupp} simplifies to:

\begin{equation}
\begin{split}
    &\sum_{t=1}^{T-1} \frac{t}{T} ~ \alpha^{t} \geq \underset{s \sim p^{\pi}_{0}}{\mathrm{Var}} ~ \bigl[\Upsilon_s\bigr] + \sum_{t=1}^{T-1} \frac{2T-3t}{T} ~ \alpha^{t}_{e} \\
    & \equiv \sum_{t=1}^{T-1} \frac{t}{T} ~ \alpha^{t} \geq \underset{s \sim p^{\pi}_{0}}{\mathrm{Var}} ~ \bigl[\Upsilon_s\bigr] + 2 \sum_{t=1}^{T-1} \frac{T-t}{T} ~ \alpha^{t}_{e} - \sum_{t=1}^{T-1} \frac{t}{T} ~ \alpha^{t}_{e} \\
    & \equiv \sum_{t=1}^{T-1} \frac{t}{T} ~ \bigl( \alpha^{t} + \alpha^{t}_{e} \bigr) \geq \underset{s \sim p^{\pi}_{0}}{\mathrm{Var}} ~ \bigl[\Upsilon_s\bigr] + 2 \sum_{t=1}^{T-1} \frac{T-t}{T} ~ \alpha^{t}_{e} \\
    & \equiv \sum_{t=1}^{T-1} \frac{t}{T} ~ \mathrm{C}_t \geq \underset{s \sim p^{\pi}_{0}}{\mathrm{Var}} ~ \bigl[\Upsilon_s\bigr] + 2 \sum_{t=1}^{T-1} \frac{T-t}{T} ~ \alpha^{t}_{e} \\
    & \equiv \sum_{t=1}^{T-1} \frac{t}{T} ~ \underset{s_t, a_t \sim p^{\pi}_{t}, \pi}{\mathrm{Cov}} ~ \bigl[ \Upsilon_{s,a}, \Upsilon_{s,a}^{t} \bigr] \geq \underset{s \sim p^{\pi}_{0}}{\mathrm{Var}} ~ \bigl[\Upsilon_s\bigr] + 2 \sum_{t=1}^{T-1} \frac{T-t}{T} ~ \underset{s_t, a_t \sim p^{\pi}_{t}, \pi}{\mathrm{Cov}} \bigl[\Upsilon_{s}, \Upsilon_{s}^{t}\bigr] \\
    & \equiv \sum_{t=1}^{T-1} \frac{t}{T} ~ \underset{s_t, a_t \sim p^{\pi}_{t}, \pi}{\mathrm{Cov}} ~ \bigl[ \Upsilon_{s,a}, \Upsilon_{s,a}^{t} \bigr] \geq \underset{s \sim p^{\pi}_{0}}{\mathrm{Var}} ~ \bigl[\Upsilon_s\bigr] + 2 \sum_{t=1}^{T-1} \frac{T-t}{T} ~ \underset{s_t, a_t \sim p^{\pi}_{t}, \pi}{\mathrm{Cov}} \bigl[\Upsilon_{s}, \Upsilon_{s}^{t}\bigr]
\end{split}
\end{equation}

Which concludes the derivation of Theorem \ref{theorem1}. 

\subsection{Derivations - Bias}
\label{appendix14}
First, we calculate the bias associated with MA (ie. MBMA and QMA), which stems from approximated state-action Q-value. We denote the approximated Q-value as $\hat{Q}^{\pi}(s,a)$ and write:

\begin{equation}
\begin{split}
    \mathrm{bias}^{MA} & = \nabla J(s,a) - \nabla \hat{J} (s,a) \\
    & = \nabla \log \pi (a|s) ~ Q^{\pi}(s,a) - \nabla \log \pi (a|s) ~ \hat{Q}^{\pi}(s,a) \\
    & = \nabla \log \pi (a|s) \bigl( Q^{\pi}(s,a) - \hat{Q}^{\pi}(s,a) \bigr)
\end{split}
\end{equation}

Furthermore, we calculate the bias associated with using dynamics models to simulate state samples. Firstly, we denote the result of a $n$-step transition via the dynamics model as $s^{*}$, such that the absolute difference between true transition and dynamics model transition is equal to $|s - s^{*}|$. Furthermore, we denote the Lipschitz norm of $\nabla \log \pi (a|s)$ as $\mathcal{K}$. As such, it follows that:

\begin{equation*}
    |\nabla \log \pi (a|s) - \nabla \log \pi (a|s^{*})| \leq \mathcal{K} |s - s^{*}|
\end{equation*}

We write the bias:

\begin{equation*}
\begin{split}
    \mathrm{bias}^{MS} & = \nabla J(s,a) - \nabla \hat{J} (s,a) \\
    & = \nabla \log \pi (a|s) ~ Q^{\pi}(s,a) - \nabla \log \pi (a|s^{*}) ~ \hat{Q}^{\pi}(s^{*},a) \\
    & = \bigl( \nabla \log \pi (a|s) - \nabla \log \pi (a|s^{*}) \bigr) \hat{Q}^{\pi}(s^{*},a) + \nabla \log \pi (a|s) \bigl(Q^{\pi}(s,a) - \hat{Q}^{\pi}(s^{*},a)\bigr) 
\end{split}
\end{equation*}

We use the Lipschitz continuity:

\begin{equation*}
    \biggl| \frac{\mathrm{bias}^{MS} - \nabla \log \pi (a|s) \bigl(Q^{\pi}(s,a) - \hat{Q}^{\pi}(s^{*},a)\bigr) }{\hat{Q}^{\pi}(s^{*},a)}\biggr| \leq \mathcal{K} |s - s^{*}|
\end{equation*}

Where we assume that $\hat{Q}^{\pi}(s^{*},a) \neq 0$. Squaring both sides leads to the solution:

\begin{equation}
\begin{split}
    & \mathrm{bias}^{MS} \geq \nabla \log \pi (a|s) \bigl( Q^{\pi}(s,a) - \hat{Q}^{\pi}(s,a) \bigr) - \sqrt{\nabla \log \pi (a|s)^2 \bigl( Q^{\pi}(s,a)^2 - Q^{\pi}(s,a)\bigr) + \bigl(\mathcal{K} (s - s^{*})\bigr)^{2}} \\
 \text{And:}\\
    & \mathrm{bias}^{MS} \leq \nabla \log \pi (a|s) \bigl( Q^{\pi}(s,a) - \hat{Q}^{\pi}(s,a) \bigr) + \sqrt{\nabla \log \pi (a|s)^2 \bigl( Q^{\pi}(s,a)^2 - Q^{\pi}(s,a)\bigr) + \bigl(\mathcal{K} (s - s^{*})\bigr)^{2}}
\end{split}
\end{equation}

Which concludes the derivation.

\section{Experimental Details}
\label{appendix3}

\subsection{Setting}
\label{appendix_setting}

\paragraph{Figure \ref{fig1}} We use the OpenAI gym CartPole environment. We define solving the environment as reaching an average of 190 rewards during $25$ evaluations. We perform policy evaluations every 50 environment steps. If the trajectory length is shorter than environment termination we bootstrap the Q-value with critic. To sample more actions per state we perform environment rewinding. Similarly to regular actions, the Q-values of additional action samples are bootstrapped via critic when reaching the trajectory length. Note that CartPole environment has only two actions, as such there is minimal variance associated with the policy. We smoothen the results with Savitsky-Golay filter and use $45$ random seeds. 

\paragraph{Table \ref{maasdc_grid2}} We use a subset of environments from DM Control Suite. We marginalize Q-values by performing $100$ rollouts for every state-action pair. We get $125 000$ on-policy states, with one additional action per state. We use Equation \ref{eq:sum} and Lemma \ref{lem:lemma1} to isolate the variance components. Note that Q-value marginalization is required by Lemma \ref{lem:lemma1}. Note that if Q-values are stochastic, we observe more variance reduction stemming from sampling additional actions than expected. The performance of agents was measured during $500 000$ environment steps, with an average performance recorded in $122$ different episodes. Additional action sample is drawn from the environments (via environment rewinding). To reduce the compute load used in the experiment, the performance is measured without Q-value marginalization. We use $10$ random seeds.

\paragraph{Table \ref{maasdc_grid00}} To measure performance we first average across random seeds and take the maximum. We normalize by dividing each seed by maximum best performing PPO seed. To measure bias and variance, we record $125$ gradient estimates for every method during $10$ points in training for $15$ random seeds. Each of $125$ gradient estimates is calculated using a batch size of $2500$ states. The gradients are always calculated wrt. the same policy. To this end, there is one agent gathering the data and serving as the policy for all methods. The recorded gradients stemming from all methods are never applied to the actor network (ie. using one agent per random seed). We denote $*$ as the tested method and $P$ as the total number of parameters in the model. We calculate relative bias with the following equation:

\begin{equation*}
    \mathrm{Bias}^{*} = \frac{1}{P} \sum_{p}^{P} \frac{|\nabla J_{p}^{*} - \nabla J_{p}^{AC}|}{|\nabla J_{p}^{*}|}    
\end{equation*}

Where $\nabla J_{p}^{*}$ and $\nabla J_{p}^{AC}$ denote the gradient wrt. $p^{th}$ parameter calculated via the tested method and actor-critic respectively (averaged over $125$ gradient examples). As such, at each testing point, we calculate the absolute difference between the 'oracle' AC gradient (which is unbiased) and the respective method average. Furthermore, we calculate relative variance via:

\begin{equation*}
    \mathrm{Var}^{*} = \frac{1}{P} \sum_{p}^{P} \frac{\underset{\tau}{\mathrm{Var}} \bigl[\nabla J_{p}^{*} \bigr]}{(\nabla J_{p}^{*})^{2}}
\end{equation*}

Where $\mathrm{Var}_{\tau} \bigl[\nabla J_{p}^{*} \bigr]$ denotes the $p^{th}$ unit of the diagonal of variance-covariance matrix calculated over $125$ gradient examples. Dividing bias and variance by the size of the gradient allows us to inspect the relative size (ie. if the gradient is small then bias and variance might also be small, but big in comparison to the gradient that we are looking for).


\subsection{Hyperparameters}

Below, we provide a detailed list of hyperparameter settings used to generate results presented in Table \ref{maasdc_grid00}.

\begin{table}[ht!]
\centering
 \begin{tabular}{||c | c c c c||} 
 \hline
 \textsc{hyperparameter} & \textsc{PPO} & \textsc{QMA} & \textsc{MBMA} & \textsc{MBPO} \\
 \hline \hline
 \textsc{action repeat} & $4$ & $4$ & $4$ & $4$ \\
 \textsc{actor optimizer} & \textsc{Adam} & \textsc{Adam} & \textsc{Adam} & \textsc{Adam} \\
 \textsc{critic optimizer} & \textsc{Adam} & \textsc{Adam} & \textsc{Adam} & \textsc{Adam} \\
 \textsc{dynamics optimizer} & --- & --- & \textsc{Adam} & \textsc{Adam} \\
 \textsc{Q-net optimizer} & --- & \textsc{Adam} & --- & --- \\
 \textsc{actor learning rate} & $3e-4$ & $3e-4$ & $3e-4$ & $3e-4$ \\
 \textsc{critic learning rate} & $3e-4$ & $3e-4$ & $3e-4$ & $3e-4$ \\
 \textsc{dynamics learning rate} & --- & --- & $3e-4$ & $3e-4$ \\
 \textsc{Q-net learning rate} & --- & $3e-4$ & --- & --- \\
 \textsc{actor optimizer epsilon} & $1e-5$ & $1e-5$ & $1e-5$ & $1e-5$ \\
 \textsc{critic optimizer epsilon} & $1e-5$ & $1e-5$ & $1e-5$ & $1e-5$ \\
 \textsc{dynamics optimizer epsilon} & --- & --- & $1e-5$ & $1e-5$ \\
 \textsc{Q-net optimizer epsilon} & --- & $1e-5$ & --- & --- \\
 \textsc{actor hidden layer size} & $512$ & $512$ & $512$ & $512$ \\
 \textsc{critic hidden layer size} & $1024$ & $1024$ & $1024$ & $1024$ \\
 \textsc{dynamics hidden layer size} & --- & --- & $1024$ & $1024$ \\
 \textsc{Q-network hidden layer size} & --- & $1024$ & --- & --- \\
 $\lambda$ & $0.95$ & $0.95$ & $0.95$ & $0.95$ \\
 \textsc{discount rate} & $0.99$ & $0.99$ & $0.99$ & $0.99$ \\
 \textsc{batch size (T)} & $2048$ & $2048$ & $2048$ & $2048$ \\
 \textsc{minibatch size} & $64$ & $64$ & $64$ & $64$ \\
 \textsc{PPO epochs} & $10$ & $10$ & $10$ & $10$ \\
 \textsc{dynamics buffer size} & --- & --- & $25000$  & $25000$ \\
 \textsc{dynamics batch size} & --- & --- & $128$ & $128$ \\
 \textsc{number of simulated actions per state (T*)} & --- & $8$ & $8$ & --- \\
\textsc{number of simulated states per state (N)} & --- & --- & --- & $8$ \\
 \textsc{simulation horizon} & --- & --- & $12$ & $12$ \\
 \textsc{clip coefficient} & $0.2$ & $0.2$ & $0.2$ & $0.2$ \\
 \textsc{maximum gradient norm} & $0.5$ & $0.5$ & $0.5$ & $0.5$ \\
 \textsc{value coefficient} & $0.5$ & $0.5$ & $0.5$ & $0.5$ \\
 \hline\hline
 \multicolumn{5}{||c||}{\textsc{quadruped and humanoid}} \\
 \hline \hline
 \textsc{batch size (T)} & $4096$ & \textsc{NA} & $4096$ & $4096$ \\
 \textsc{minibatch size} & $128$ & \textsc{NA} & $128$ & $128$ \\
 \textsc{dynamics buffer size} & --- & \textsc{NA} & $2000000$  & $2000000$ \\
 \textsc{dynamics batch size} & --- & \textsc{NA} & $256$ & $256$ \\
 \hline
\end{tabular}
\label{maasdc_grid}
\end{table}

\subsection{Computational Costs}

Below, we report the relative computational costs associated with each SPG update type. Note, that code optimization and parallelization would increase the relative performance of QMA and MBMA.

\begin{table}[ht!]
\centering
 \begin{tabular}{||c | c c c||} 
 \hline
 \textsc{Number of actions} & \textsc{PPO} & \textsc{QMA} & \textsc{MBMA} \\
 \hline \hline
 \textsc{4} & $1.00$ & $1.22$ & $1.59$ \\
 \textsc{8} & $1.00$ & $1.62$ & $2.31$ \\
 \textsc{16} & $1.00$ & $2.07$ & $3.60$ \\
 \hline
\end{tabular}
\vspace{-0.1in}
\end{table}

\newpage

\subsection{Unnormalized Results}
\label{appendix24}

We provide a table of unnormalized results for the performance experiment:

\begin{table}[h]
\centering
\vskip 0.15in
 {\renewcommand{\arraystretch}{1.1}%
\begin{tabular}{||l||c|c|c|c||}
\hline
    \textsc{Task} & \textsc{PPO} & \textsc{MBMA} & \textsc{MBPO} & \textsc{QMA} \\
    \hline\hline    
    \textsc{acro swingup} & $47\pm10 \, (32\pm7)$ & $\mathbf{59\pm6 \, (40\pm6)}$ & $56\pm8 \, (31\pm7)$ & $34\pm10 \, (17\pm7)$ \\
    \textsc{ball catch} & $948\pm8 \, (831\pm20)$ & $\mathbf{974\pm3 \, (898\pm8)}$ & $969\pm2 \, (888\pm8)$ & $934\pm5 \, (770\pm30)$ \\   
    \textsc{cart swingup} & $736\pm46 \, (617\pm46)$ & $\mathbf{828\pm16 \, (707\pm44)}$ & $825\pm13 \, (702\pm38)$ & $802\pm2 \, (677\pm18)$ \\
    \textsc{cart 2-pole} & $308\pm16 \, (253\pm8)$ & $\mathbf{575\pm52 \,(388\pm27)}$ & $435\pm48 \, (317\pm22)$ & $315\pm22 \, (248\pm12)$ \\
    \textsc{cart 3-pole} & $229\pm15 \, (199\pm10)$ & $229\pm14 \, (202\pm10)$ & $261\pm6 \, \mathbf{(221\pm9)}$ & $\mathbf{262\pm5} \, (211\pm4)$ \\
    \textsc{cheetah run} & $283\pm12 \, (185\pm10)$ & $\mathbf{507\pm14 \, (316\pm14)}$ & $473\pm17 \, (284\pm14)$ & $201\pm8 \, (135\pm7)$ \\
    \textsc{finger spin} & $\mathbf{391\pm21 \, (280\pm14)}$ & $350\pm14 \, (266\pm12)$ & $305\pm16 \, (248\pm14)$ & $359\pm15 \, (245\pm12)$ \\
    \textsc{finger turn} & $\mathbf{396\pm67 \, (213\pm54)}$ & $368\pm59 \, (206\pm52)$ & $318\pm73 \, (184\pm51)$ & $296\pm75 \, (176\pm50)$ \\
    \textsc{point easy} & $895\pm6 \, (839\pm13)$ & $\mathbf{910\pm5} \, (866\pm7)$ & $909\pm6 \, \mathbf{(867\pm7)}$ & $467\pm97 \, (106\pm50)$ \\
    \textsc{reacher easy} & $885\pm44 \, (649\pm66)$ & $\mathbf{968\pm2 \, (815\pm39)}$ & $854\pm71 \, (729\pm74)$ & $472\pm27 \, (316\pm72)$ \\
    \textsc{reacher hard} & $601\pm103 \, (385\pm78)$ & $767\pm96 \, (606\pm84)$ & $\mathbf{892\pm61 \, (722\pm61)}$ & $488\pm53 \, (361\pm47)$ \\
    \textsc{walker stand} & $914\pm22 \, (737\pm24)$ & $\mathbf{955\pm5 \, (839\pm22)}$ & $944\pm8 \, (815\pm29)$ & $854\pm20 \, (654\pm21)$ \\
    \textsc{walker walk} & $514\pm14 \, (377\pm14)$ & $\mathbf{892\pm9 \, (686\pm18)}$ & $720\pm19 \, (576\pm19)$ & $500\pm17 \, (340\pm14)$ \\
    \textsc{walker run} & $203\pm7 \, (152\pm5)$ & $\mathbf{331\pm13 \, (251\pm9)}$ & $233\pm12 \, (190\pm11)$ & $208\pm7 \, (141\pm4)$ \\
    \hline\hline
\end{tabular}}
\label{maasdc_grid9}
\end{table}

And for the bias-variance experiment:

\begin{table}[h]
\centering
\vskip 0.1in
 {\renewcommand{\arraystretch}{1.1}%
\begin{tabular}{||l||c|c|c||c|c|c|c||}
\hline
    & \multicolumn{3}{c||}{\textsc{Relative Bias}} & \multicolumn{4}{c||}{\textsc{Relative Variance}} \\  \hline
    \textsc{Task} & \textsc{MBMA} & \textsc{MBPO} & \textsc{QMA} & \textsc{AC} & \textsc{MBMA} & \textsc{MBPO} & \textsc{QMA} \\
    \hline\hline
    \textsc{acro swingup} & $\mathbf{8.25\pm0.3}$ & $8.66\pm0.3$ & $10.5\pm0.5$ & $44.3\pm1.5$ & $31.6\pm1.1$ & $30.8\pm1.0$ & $\mathbf{27.3\pm1.1}$ \\
    \textsc{ball catch} & $\mathbf{5.28\pm0.3} $& $6.04\pm0.3$ & $12.1\pm0.8$ & $48.7\pm1.4$ & $20.0\pm0.9$ & $\mathbf{18.0\pm0.7}$ & $18.8\pm1.1$ \\
    \textsc{cart swingup} & $\mathbf{3.16\pm0.3}$ & $3.46\pm0.3$ & $8.89\pm1.1$ & $21.2\pm1.5$ & $8.08\pm0.6$ & $\mathbf{7.84\pm0.6}$ & $11.1\pm1.1$ \\
    \textsc{cart 2-pole} & $\mathbf{6.32\pm0.4}$ & $6.79\pm0.5$ & $10.6\pm0.7$ & $39.8\pm1.8$ & $21.6\pm1.6$ & $\mathbf{20.5\pm1.6}$ & $29.8\pm1.4$ \\
    \textsc{cart 3-pole} & $7.48\pm0.7$ & $\mathbf{7.29\pm0.6}$ & $11.0\pm0.7$ & $43.0\pm2.7$ & $\mathbf{27.7\pm2.7}$ & $29.6\pm2.8$ & $32.4\pm2.3$ \\
    \textsc{cheetah run} & $\mathbf{11.5\pm0.4}$ & $12.0\pm0.4$ & $21.4\pm0.8$ & $77.1\pm2.5$ & $39.0\pm1.1$ & $\mathbf{37.7\pm1.1}$ & $52.9\pm1.7$ \\
    \textsc{finger spin} & $\mathbf{4.50\pm0.3}$ & $4.91\pm0.4$ & $9.60\pm0.4$ & $38.8\pm1.1$ & $\mathbf{24.2\pm1.3}$ & $33.6\pm2.0$ & $28.2\pm0.9$ \\
    \textsc{finger turn} & $\mathbf{3.55\pm0.4}$ & $3.94\pm0.4$ & $11.3\pm0.8$ & $51.0\pm2.8$ & $\mathbf{25.4\pm1.8}$ & $30.9\pm2.7$ & $30.4\pm2.1$ \\
    \textsc{point easy} & $\mathbf{1.33\pm0.1}$ & $1.49\pm0.1$ & $3.57\pm0.6$ & $15.8\pm1.4$ & $4.40\pm0.4$ & $\mathbf{3.84\pm0.3}$ & $4.08\pm0.5$ \\
    \textsc{reacher easy} & $\mathbf{4.07\pm0.2}$ & $4.85\pm0.3$ & $10.3\pm1.0$ & $36.2\pm1.8$ & $14.3\pm0.7$ & $\mathbf{13.8\pm0.7}$ & $15.6\pm1.4$ \\
    \textsc{reacher hard} & $\mathbf{5.70\pm0.3}$ & $6.58\pm0.4$ & $13.0\pm0.7$ & $41.6\pm1.4$ & $21.1\pm1.2$ & $\mathbf{20.0\pm1.1}$ & $23.1\pm1.4$ \\
    \textsc{walker stand} & $\mathbf{9.77\pm0.4}$ & $11.4\pm0.5$ & $18.8\pm0.8$ & $71.2\pm2.0$ & $\mathbf{39.9\pm1.2}$ & $43.0\pm1.4$ & $51.5\pm2.1$ \\
    \textsc{walker walk} & $\mathbf{11.1\pm0.4}$ & $12.3\pm0.4$ & $16.6\pm0.8$ & $76.7\pm1.8$ & $\mathbf{40.7\pm1.2}$ & $42.0\pm1.2$ & $59.4\pm1.1$ \\
    \textsc{walker run} & $\mathbf{11.1\pm0.4}$ & $12.1\pm0.4$ & $15.7\pm0.8$ & $77.9\pm1.8$ & $ 41.3\pm1.3 $ & $\mathbf{40.7\pm1.0}$ & $59.5\pm1.6$ \\
    \hline\hline
\end{tabular}}
\label{maasdc_grid007}
\vskip -0.1in
\end{table}

\newpage

\section{Learning Curves}
\label{appendix22}

\begin{figure}[ht!]
\label{fig:testq}
\centering
\vspace{5pt}
\begin{subfigure}{.3\linewidth}
    \includegraphics[width=\textwidth]{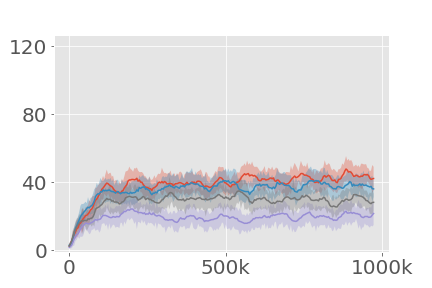}
    \caption{acrobot swingup}
\end{subfigure}
\begin{subfigure}{.3\linewidth}
    \includegraphics[width=\textwidth]{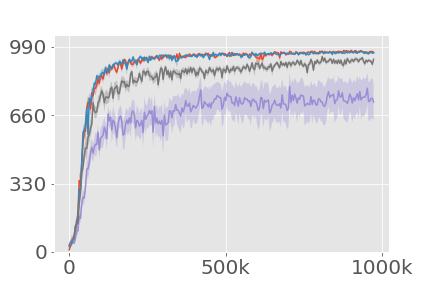}
    \caption{ball catch}
\end{subfigure}
\begin{subfigure}{.3\linewidth}
    \includegraphics[width=\textwidth]{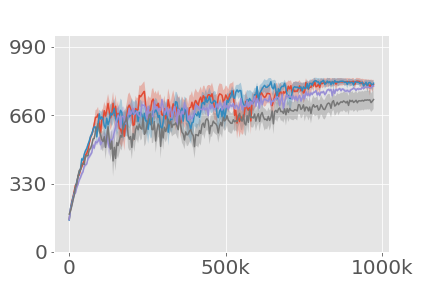}
    \caption{cartpole swingup}
\end{subfigure}
\smallskip
\begin{subfigure}{.3\linewidth}
    \includegraphics[width=\textwidth]{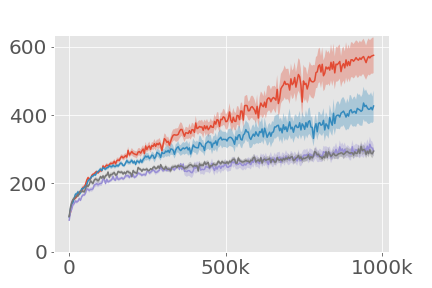}
    \caption{cartpole 2-poles}
\end{subfigure}
\begin{subfigure}{.3\linewidth}
    \includegraphics[width=\textwidth]{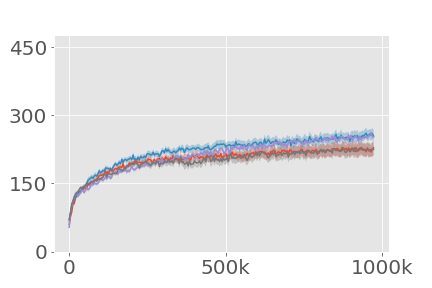}
    \caption{cartpole 3-poles}
\end{subfigure}
\begin{subfigure}{.3\linewidth}
    \includegraphics[width=\textwidth]{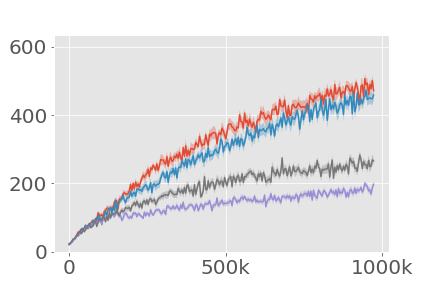}
    \caption{cheetah run}
\end{subfigure}
\smallskip
\begin{subfigure}{.3\linewidth}
    \includegraphics[width=\textwidth]{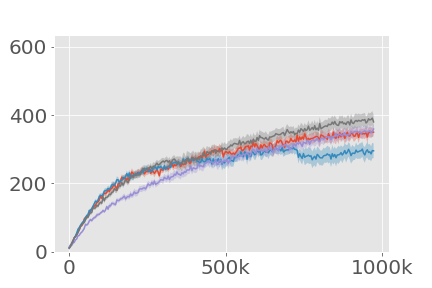}
    \caption{finger spin}
\end{subfigure}
\begin{subfigure}{.3\linewidth}
    \includegraphics[width=\textwidth]{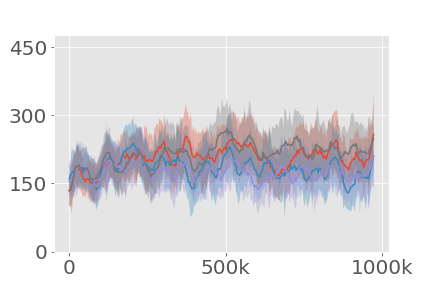}
    \caption{finger turn easy}
\end{subfigure}
\begin{subfigure}{.3\linewidth}
    \includegraphics[width=\textwidth]{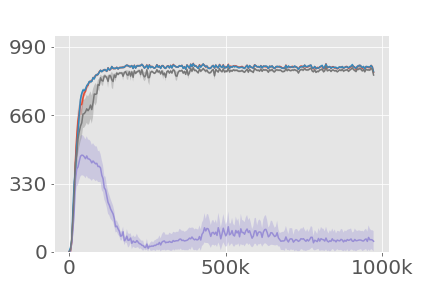}
    \caption{point-mass easy}
\end{subfigure}
\smallskip
\begin{subfigure}{.3\linewidth}
    \includegraphics[width=\textwidth]{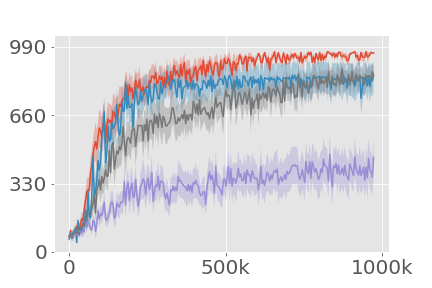}
    \caption{reacher easy}
\end{subfigure}
\begin{subfigure}{.3\linewidth}
    \includegraphics[width=\textwidth]{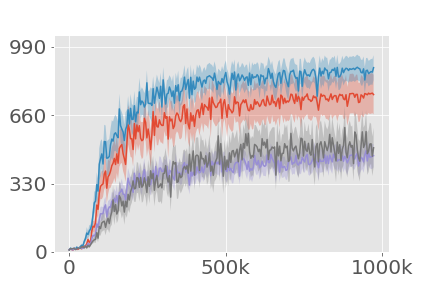}
    \caption{reacher hard}
\end{subfigure}
\begin{subfigure}{.3\linewidth}
    \includegraphics[width=\textwidth]{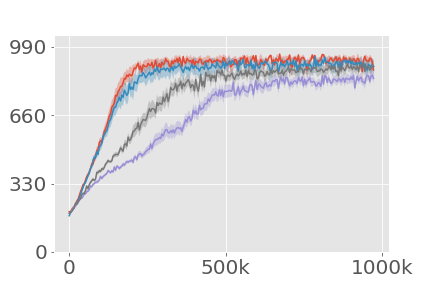}
    \caption{walker stand}
\end{subfigure}
\smallskip
\begin{subfigure}{.3\linewidth}
    \includegraphics[width=\textwidth]{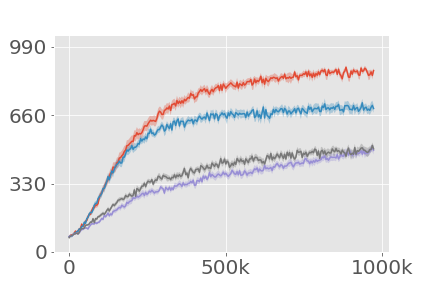}
    \caption{walker walk}
\end{subfigure}
\begin{subfigure}{.3\linewidth}
    \includegraphics[width=\textwidth]{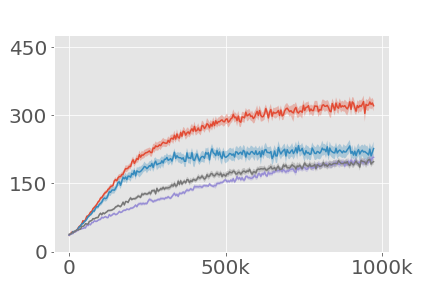}
    \caption{walker run}
\end{subfigure}
\begin{subfigure}{.3\linewidth}
    \includegraphics[width=\textwidth]{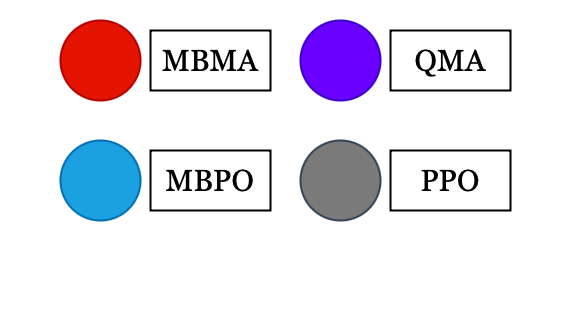}
    \caption{legend}
\end{subfigure}
\caption{Learning curves corresponding to results from Table \ref{maasdc_grid00}. The shaded region denotes one standard deviation of the mean. 15 seeds.}
\end{figure}

\section{Ablations}
\label{appendix21}

We investigate the performance of MBMA and MBPO for different $N$ and $T$ (amount of simulated samples and simulation horizon). We record the average final performance during training of $500k$ steps. First, we investigate the effects of $N$. The table below shows the mean performance for 5 DMC tasks and various amount of simulated data ($10$ seeds):

\begin{table}[ht!]
\centering
 \begin{tabular}{||c | c c c c||} 
 \hline
 \textsc{Method} & \textsc{N} = 8 & \textsc{N} = 16 & \textsc{N} = 32 & \textsc{N} = 64 \\
 \hline \hline
 \textsc{QMA} & $\mathbf{551}$ & $390$ & $303$ & $227$ \\
 \textsc{MBPO} & $\mathbf{636}$ & $624$ & $570$ & $594$ \\
 \textsc{MBMA} & $683$ & $641$ & $679$ & $\mathbf{707}$ \\
 \hline
\end{tabular}
\end{table}

As the table below shows, MBMA is much more robust to the amount of simulated data. Furthermore, we investigate the effects of different simulation horizons. The table below shows the mean performance for 5 DMC tasks and various simulation lengths ($10$ seeds):

\begin{table}[ht!]
\centering
 \begin{tabular}{||c | c c c||} 
 \hline
 \textsc{Method} & \textsc{H} = 12 & \textsc{H} = 24 & \textsc{H} = 48 \\
 \hline \hline
 \textsc{MBPO} & $\mathbf{636}$ & $592$ & $542$ \\
 \textsc{MBMA} & $\mathbf{683}$ & $604$ & $551$ \\
 \hline
\end{tabular}
\end{table}

We find that MBMA is more robust to hyperparameter settings than MBPO. As discussed in the main body of the paper, this most likely stems from the more favorable bias variance structure of MBMA.

\section{MBMA Implementation Details}
\label{newappendix}

We implement MBMA on top of PPO implementation \citep{schulman2017proximal} taken from CleanRL repository \citep{huang2022cleanrl}. Besides actor and critic networks which are standard for PPO, MBMA uses a dynamics model. Following \citet{janner2019trust}, we implement a simplistic dynamics model consisting of reward and transition models working directly on proprioceptive state representations. 

\paragraph{Reward model} inputs concatenated state-action and outputs scalar value per state-action. It is trained using MSE loss function with real reward used as a target.

\paragraph{Transition model} inputs concatenated state-action and outputs a state vector per state-action pair. It is trained using MSE loss function with future state used as a target. 

\paragraph{Critic} inputs state and outputs scalar value per state. It is trained using MSE loss function with discounted sum of rollout rewards (ie. Monte Carlo) used as a target.

\paragraph{Actor} inputs state and outputs means of a Gaussian distribution. It is trained using PPO clipped objective using a mix of real on-policy data (generated exactly as a regular implementation of PPO would) and simulated on-policy data (which consists of Q-values of additional actions sampled at real on-policy states). The simulated Q-values are calculated by unrolling the dynamics model for some number of steps ("simulation horizon" hyperparameter) and bootstrapping it with future state value given by the critic. 


\end{document}